# Description Logic Knowledge and Action Bases

**Babak Bagheri Hariri**                                 BAGHERI@INF.UNIBZ.IT
**Diego Calvanese**                                     CALVANESE@INF.UNIBZ.IT
**Marco Montali**                                        MONTALI@INF.UNIBZ.IT
*KRDB Research Centre for Knowledge and Data*
*Free University of Bozen-Bolzano*
*Piazza Domenicani 3, 39100 Bolzano, Italy*

**Giuseppe De Giacomo**                               DEGIACOMO@DIS.UNIROMA1.IT
**Riccardo De Masellis**                              DEMASELLIS@DIS.UNIROMA1.IT
**Paolo Felli**                                          FELLI@DIS.UNIROMA1.IT
*Dipartimento di Ingegneria Informatica Automatica e Gestionale*
*Sapienza Università di Roma*
*Via Ariosto 25, 00185 Roma, Italy*

## Abstract

Description logic Knowledge and Action Bases (KAB) are a mechanism for providing both a semantically rich representation of the information on the domain of interest in terms of a description logic knowledge base and actions to change such information over time, possibly introducing new objects. We resort to a variant of *DL-Lite* where the unique name assumption is not enforced and where equality between objects may be asserted and inferred. Actions are specified as sets of conditional effects, where conditions are based on epistemic queries over the knowledge base (TBox and ABox), and effects are expressed in terms of new ABoxes. In this setting, we address verification of temporal properties expressed in a variant of first-order $\mu$-calculus with quantification across states. Notably, we show decidability of verification, under a suitable restriction inspired by the notion of weak acyclicity in data exchange.

## 1. Introduction

Recent work in business processes, services and databases is bringing forward the need of considering both data and processes as first-class citizens in process and service design (Nigam & Caswell, 2003; Bhattacharya, Gerede, Hull, Liu, & Su, 2007; Deutsch, Hull, Patrizi, & Vianu, 2009; Vianu, 2009; Meyer, Smirnov, & Weske, 2011). In particular, the so-called artifact-centric approaches, which advocate a sort of middle ground between a conceptual formalization of dynamic systems and their actual implementation, are promising to be effective in practice (Cohn & Hull, 2009). The verification of temporal properties in the presence of data represents a significant research challenge (for a survey, see Calvanese, De Giacomo, & Montali, 2013), since taking into account how data evolve over time results in systems that have an infinite number of states. Neither finite-state model checking (Clarke, Grumberg, & Peled, 1999) nor most of the current techniques for infinite-state model checking, which mostly tackle recursion (Burkart, Caucal, Moller, & Steffen, 2001), apply to this case. Recently, there have been some advancements on this issue (Cangialosi, De Giacomo, De Masellis, & Rosati, 2010; Damaggio, Deutsch, & Vianu, 2011; Bagheri Hariri, Calvanese, De Giacomo, De Masellis, & Felli, 2011; Belardinelli, Lomuscio, & Patrizi, 2011), in the context of suitably constrained relational database settings.

While most of this work is based on maintaining information in a relational database, for more sophisticated applications it is foreseen to enrich data-intensive business processes with a semantic level, where information can be maintained in a semantically rich knowledge base which allows for operating with incomplete information (Calvanese, De Giacomo, Lembo, Montali, & Santoso, 2012; Limonad, De Leenheer, Linehan, Hull, & Vaculin, 2012). This leads us to look into how to combine first-order data, ontologies, and processes, while maintaining basic inference tasks (specifically verification) decidable. In this setting, we capture the domain of interest in terms of semantically rich formalisms as those provided by ontological languages based on Description Logics (DLs) (Baader, Calvanese, McGuinness, Nardi, & Patel-Schneider, 2003). Such languages natively deal with incomplete knowledge in the modeled domain. This additional flexibility comes with an added cost, however: differently from relational databases, to evaluate queries we need to resort to logical implication. Moreover, incomplete information combined with the ability of evolving the system through actions results in a notoriously fragile setting w.r.t. decidability (Wolter & Zakharyaschev, 1999b, 1999a; Gabbay, Kurusz, Wolter, & Zakharyaschev, 2003). In particular, due to the nature of DL assertions (which in general are not definitions but constraints on models), we get one of the most difficult kinds of domain descriptions for reasoning about actions (Reiter, 2001), which amounts to dealing with complex forms of state constraints (Lin & Reiter, 1994).

To overcome this difficulty, virtually all solutions that aim at robustness are based on a so-called "functional view of knowledge bases" (Levesque, 1984): the KB provides the ability of querying based on logical implication ("ask"), and the ability of progressing it to a "new" KB through forms of updates ("tell") (Baader, Ghilardi, & Lutz, 2012; Calvanese, De Giacomo, Lenzerini, & Rosati, 2011). Notice that this functional view is tightly related to an epistemic interpretation of the KB (Calvanese, De Giacomo, Lembo, Lenzerini, & Rosati, 2007a). Indeed our work is also related to that on Epistemic Dynamic Logic (van Ditmarsch, van der Hoek, & Kooi, 2007), and, though out of the scope of this paper, the decidability results presented here could find application in the context of that research as well.

We follow this functional view of KBs. However, a key point of our work is that at each execution step external information is incorporated into the system in form of new individuals (denoted by *function terms*), that is, our systems are not closed w.r.t. the available information. This makes our framework particularly interesting and challenging. In particular, the presence of these individuals requires a specific treatment of equality, since as the system progresses and new information is acquired, distinct function terms may be inferred to denote the same object.

Specifically, we introduce the so-called *Knowledge and Action Bases* (KABs). A KAB is equipped with an ontology or, more precisely, a TBox expressed, in our case, in a variant of *DL-Lite$_\mathcal{A}$* (Calvanese, De Giacomo, Lembo, Lenzerini, & Rosati, 2007b), which extends the core of the Web Ontology Language OWL 2 QL (Motik, Cuenca Grau, Horrocks, Wu, Fokoue, & Lutz, 2012) and is particularly well suited for data management. Such a TBox captures intensional information on the domain of interest, similarly to UML class diagrams or other conceptual data models, though as a software component to be used at run-time. The KAB includes also an ABox, which acts as a storage or state. The ABox maintains the data of interest, which are accessed by relying on query answering based on logical implication (certain answers). Notably, our variant of *DL-Lite$_\mathcal{A}$* is without the unique name assumption (UNA), and we allow for explicit equality assertions in the ABox. In this way we can suitably treat function terms to represent individuals acquired during the execution. Technically, the need of dealing with equality breaks the first-order rewritability of *DL-Lite$_\mathcal{A}$* query answering, and requires that, in addition to the rewriting process, inference on equality is performed (Artale, Calvanese, Kontchakov, & Zakharyaschev, 2009). As a query language, we use unions of conjunctive queries, possibly composing their certain answers through full FOL constructs. This gives rise to an *epistemic query language* that asks about what is "known" by the current KB (Calvanese et al., 2007a). Apart from the KB, the KAB contains *actions*, whose execution changes the state of the KB, i.e., its ABox. Such actions are specified as sets of *conditional effects*, where conditions are (epistemic) queries over the KB and effects are expressed in terms of new ABoxes. Actions have no static pre-conditions, whereas a process is used to specify which actions can be executed at each step. For simplicity, we model such processes as condition/action rules, where the condition is again expressed as a query over the KB.

In this setting, we address the verification of temporal/dynamic properties expressed in a first-order variant of $\mu$-calculus (Park, 1976; Stirling, 2001), where atomic formulae are queries over the KB which can refer both to constants and to function terms, and where a controlled form of quantification across states is allowed. Notice that all previous decidability results on actions over DL KBs assumed that no information is coming from outside of the system, in the sense that no new individual terms are added while executing actions (Calvanese et al., 2011; Baader et al., 2012; Rosati & Franconi, 2012). In this paper, instead, we allow for arbitrary introduction of new terms. Unsurprisingly, we show that even for very simple KABs and temporal properties, verification is undecidable. However, we also show that for a rich class of KABs, verification is in fact decidable and reducible to finite-state model checking. To obtain this result, following Cangialosi et al. (2010), and Bagheri Hariri et al. (2011), we rely on recent results in data exchange on the finiteness of the chase of tuple-generating dependencies (Fagin, Kolaitis, Miller, & Popa, 2005), though, in our case, we need to extend the approach to deal with *(i)* incomplete information, *(ii)* inference on equality, and *(iii)* quantification across states in the verification language.

The paper is organized as follows. In Section 2 we give preliminaries about *DL-Lite$_{\mathcal{A}}$* without UNA, which is going to be our knowledge base formalism. Section 3 describes the KAB framework in detail, while Section 4 discusses its execution semantics. In Section 5 we introduce the verification formalism for KABs. In Section 6, we show that verification of KABs is in general undecidable, even considering very simple temporal properties and KABs. In Section 7, we give our main technical result: verification of weakly acyclic KABs is decidable in EXPTIME. In Section 8, we extensively survey related work. Section 9 concludes the paper.

## 2. Knowledge Base Formalism

Description Logics (DLs) (Baader et al., 2003) are knowledge representation formalisms that are tailored for representing the domain of interest in terms of *concepts* (or classes), denoting sets of objects, and *roles* (or relations), denoting binary relations between objects. DL *knowledge bases* (KBs) are based on an alphabet of concept and role names, and an alphabet of individuals. A DL KB is formed by two distinct parts: a *TBox*, which represents the *intensional level* of the KB and contains a description of the domain of interest in terms of universal assertions over concepts and roles; and an *ABox*, which represents the *instance level* of the KB and contains extensional information on the participation of individuals to concepts and roles.

For expressing KBs we use *DL-Lite$_{\text{NU}}$*, a variant of the *DL-Lite$_{\mathcal{A}}$* language (Poggi, Lembo, Calvanese, De Giacomo, Lenzerini, & Rosati, 2008; Calvanese, De Giacomo, Lembo, Lenzerini, & Rosati, 2013) in which we drop the *unique name assumption* (UNA) in line with the standard Web Ontology Language (OWL 2) (Bao et al., 2012). Essentially, *DL-Lite$_{\text{NU}}$* extends the OWL 2 QL profile of OWL 2, by including functionality assertions and the possibility to state equality between individuals.

The syntax of *concept* and role *expressions* in *DL-Lite*$_{\text{NU}}$ is as follows:

$$B \longrightarrow N \mid \exists R, \qquad R \longrightarrow P \mid P^-,$$
$$C \longrightarrow B \mid \neg B, \qquad V \longrightarrow R \mid \neg R,$$

where $N$ denotes a *concept name*, $P$ a *role name*, and $P^-$ an *inverse role*.

Formally, in a *DL-Lite*$_{\text{NU}}$ KB $(T, A)$, the TBox $T$ is a finite set of *TBox assertions* of the form

$$B \sqsubseteq C, \qquad R \sqsubseteq V, \qquad (\text{funct } R),$$

called respectively *concept inclusions*, *role inclusions*, and *functionality assertions*. We follow the usual assumption in *DL-Lite*, according to which a TBox may contain neither (funct $P$) nor (funct $P^-$) if it contains $R \sqsubseteq P$ or $R \sqsubseteq P^-$, for some role $R$ (Poggi et al., 2008; Calvanese et al., 2013). This condition expresses that roles in functionality assertions cannot be specialized.

*DL-Lite*$_{\text{NU}}$ TBoxes are able to capture the essential features of conceptual modeling formalisms, such as UML Class Diagrams (or Entity-Relationship schemas), namely ISA between classes and associations (relationships), disjointness between classes and between associations, typing of associations, and association multiplicities (in particular, mandatory participation and functionality). The main missing feature is completeness of hierarchies, which would require the introduction of disjunction and would compromise the good computational properties of *DL-Lite*.

The ABox $A$ in a *DL-Lite*$_{\text{NU}}$ KB $(T, A)$ is a finite set of *ABox assertions* of the form

$$N(t_1), \qquad P(t_1, t_2), \qquad t_1 = t_2,$$

called respectively, *concept (membership) assertions*, *role (membership) assertions*, and *equality assertions*, where $t_1, t_2$ are terms denoting individuals (see below). The presence of equality assertions in the ABox requires a specific treatment of equality that goes beyond the usual reasoning techniques for *DL-Lite* based on first-order rewritability, although reasoning remains polynomial (Artale et al., 2009). On the other hand, we do not allow for explicit disequality, though one can use membership in disjoint concepts to assert that two individuals are different.

*DL-Lite*$_{\text{NU}}$ admits complex terms for denoting individuals. Such terms are inductively defined by starting from a finite set of constants, and applying a finite set of (uninterpreted) functions of various arity greater than 0. As a result, the set of individual terms is countably infinite. We call *function terms* those terms involving functions. Also, the structure of terms has an impact on inference over equality, which is a congruence relation on the structure of terms, i.e., if $t_i = t'_i$, for $i \in \{1, \ldots, n\}$, and $f$ is a function symbol of arity $n$, then $f(t_1, \ldots, t_n) = f(t'_1, \ldots, t'_n)$. Apart from this aspect related to equality, we can treat individuals denoted by terms simply as ordinary individual constants in DLs.

We adopt the standard semantics of DLs based on FOL interpretations $\mathcal{I} = (\Delta^{\mathcal{I}}, \cdot^{\mathcal{I}})$, where $\Delta^{\mathcal{I}}$ is the interpretation domain and $\cdot^{\mathcal{I}}$ is the interpretation function such that $t^{\mathcal{I}} \in \Delta^{\mathcal{I}}$, $N^{\mathcal{I}} \subseteq \Delta^{\mathcal{I}}$, and $P^{\mathcal{I}} \subseteq \Delta^{\mathcal{I}} \times \Delta^{\mathcal{I}}$, for each term $t$, concept name $N$, and role name $P$. Coherently with the congruence relation on terms, we have that $(f(t_1, \ldots, t_n))^{\mathcal{I}} = (f(t'_1, \ldots, t'_n))^{\mathcal{I}}$, whenever $t_i^{\mathcal{I}} = t'^{\mathcal{I}}_i$, for $i \in \{1, \ldots, n\}$.

Complex concepts and roles are interpreted as follows:

$$(\exists R)^{\mathcal{I}} = \{o \mid \exists o'.(o, o') \in R^{\mathcal{I}}\}, \qquad (P^-)^{\mathcal{I}} = \{(o_1, o_2) \mid (o_2, o_1) \in P^{\mathcal{I}}\},$$
$$(\neg B)^{\mathcal{I}} = \Delta^{\mathcal{I}} \setminus B^{\mathcal{I}}, \qquad (\neg R)^{\mathcal{I}} = \Delta^{\mathcal{I}} \times \Delta^{\mathcal{I}} \setminus R^{\mathcal{I}}.$$

An interpretation $\mathcal{I}$ *satisfies* an assertion of the form:
- $B \sqsubseteq C$, if $B^{\mathcal{I}} \subseteq C^{\mathcal{I}}$;
- $R \sqsubseteq V$, if $R^{\mathcal{I}} \subseteq V^{\mathcal{I}}$;
- (funct $R$), if for all $o, o_1, o_2$ we have that, if $\{(o, o_1), (o, o_2)\} \subseteq R^{\mathcal{I}}$, then $o_1 = o_2$;
- $N(t_1)$, if $t_1^{\mathcal{I}} \in N^{\mathcal{I}}$;
- $P(t_1, t_2)$, if $(t_1^{\mathcal{I}}, t_2^{\mathcal{I}}) \in P^{\mathcal{I}}$;
- $t_1 = t_2$, if $t_1^{\mathcal{I}} = t_2^{\mathcal{I}}$.

$\mathcal{I}$ is a *model* of a KB $(T, A)$ if it satisfies all assertions in $T$ and $A$. KB $(T, A)$ is *satisfiable* if it has a model. We also say that an ABox $A$ is *consistent w.r.t. a TBox $T$* if the KB $(T, A)$ is satisfiable. An assertion $\alpha$ is *logically implied* by a KB $(T, A)$, denoted $(T, A) \models \alpha$, if every model of $(T, A)$ satisfies $\alpha$ as well.

The following characterization of satisfiability and logical implication in *DL-Lite*$_{\text{NU}}$ is an easy consequence of results by Artale et al. (2009).

**Theorem 1** *Checking satisfiability and logical implication in DL-Lite$_{\text{NU}}$ are* PTIME-*complete.*

*Proof.* The PTIME lower bound is an immediate consequence of the same lower bound established by Artale et al. (2009) for *DL-Lite*$_{\text{NU}}$ in which we do not allow the use of complex individual terms.

For the upper bound, Artale et al. (2009) provide a PTIME algorithm that is based on first using functionality assertions to exhaustively propagate equality, and then resorting to a PTIME algorithm (in combined complexity) for reasoning in *DL-Lite* in the absence of UNA. We can adapt that algorithm by changing the first step, so as to propagate, again in PTIME, equality over terms in the active domain not only due to functionalities, but also due to congruence. □

Next we introduce queries. As usual (cf. OWL 2), answers to queries are formed by constants/terms denoting individuals explicitly mentioned in the ABox. The *(active) domain of an ABox $A$*, denoted by ADOM($A$), is the (finite) set of constants/terms appearing in concept, role, and equality assertions in $A$. The (predicate) alphabet of a KB $(T, A)$, denoted ALPH($(T, A)$) is the set of concept and role names occurring in $T \cup A$.

A *union of conjunctive queries* (UCQ) $q$ over a KB $(T, A)$ is a FOL formula of the form $\exists \vec{y}_1.conj_1(\vec{x}, \vec{y}_1) \vee \cdots \vee \exists \vec{y}_n.conj_n(\vec{x}, \vec{y}_n)$ with free variables $\vec{x}$ and existentially quantified variables $\vec{y}_1, \ldots, \vec{y}_n$. Each $conj_i(\vec{x}, \vec{y}_i)$ in $q$ is a conjunction of atoms of the form $N(z), P(z, z')$ where $N$ and $P$ respectively denote a concept and a role name occurring in ALPH($(T, A)$), and $z, z'$ are constants in ADOM($A$) or variables in $\vec{x}$ or $\vec{y}_i$, for some $i \in \{1, \ldots, n\}$. The *certain answers* to $q$ over $(T, A)$ is the set ANS$(q, T, A)$ of substitutions[1] $\sigma$ of the free variables of $q$ with constants/terms in ADOM($A$) such that $q\sigma$ evaluates to true in every model of $(T, A)$, i.e., $q\sigma$ is logically implied by $(T, A)$. Following the notation used for assertions, we denote this as $(T, A) \models q\sigma$. If $q$ has no free variables, then it is called *boolean* and its certain answers are either the empty substitution denoting true or nothing denoting false.

Again, as an easy consequence of the results by Artale et al. (2009), we obtain the following characterization of query answering in *DL-Lite*$_{\text{NU}}$.

**Theorem 2** *Computing* ANS$(q, T, A)$ *of an UCQ $q$ over a DL-Lite$_{\text{NU}}$ KB $(T, A)$ is* PTIME-*complete in the size of $T$ and $A$.*

---

1. As customary, we can view each substitution simply as a tuple of constants, assuming some ordering of the free variables of $q$.

*Proof.* As in the proof of Theorem 1, we can first propagate in PTIME equality over terms in the active domain by using functionality and congruence closure, and then resort to query answering in *DL-Lite* in the presence of UNA, which is in PTIME in the combined size of the TBox $T$ and the ABox resulting from the above equality propagation (actually, in $AC^0$ in the size of this ABox). □

We also consider an extension of UCQs, called ECQs, which are queries of the query language *EQL-Lite*(UCQ) (Calvanese et al., 2007a), that is, the FOL query language whose atoms are UCQs evaluated according to the certain answer semantics above. An *ECQ* over a KB $(T, A)$ is a possibly open formula of the form

$$Q \longrightarrow [q] \mid [x = y] \mid \neg Q \mid Q_1 \wedge Q_2 \mid \exists x.Q,$$

where $[q]$ denotes the certain answers of a UCQ $q$ over $(T, A)$, $[x = y]$ denotes the certain answers of $x = y$ over $(T, A)$, that is, the set $\{\langle x, y \rangle \in \text{ADOM}(A) \mid (T, A) \models (x = y)\}$, logical operators have the usual meaning, and quantification ranges over elements of $\text{ADOM}(A)$.

Formally we define the relation $Q$ *holds in* $(T, A)$ *under substitution* $\sigma$ of all free variables in $Q$, written $T, A, \sigma \models Q$, inductively as follows:

$$\begin{array}{lll}
T, A, \sigma \models [q] & \text{if} & (T, A) \models q\sigma, \\
T, A, \sigma \models [x = y] & \text{if} & (T, A) \models (x = y)\sigma, \\
T, A, \sigma \models \neg Q & \text{if} & T, A, \sigma \not\models Q, \\
T, A, \sigma \models Q_1 \wedge Q_2 & \text{if} & T, A, \sigma \models Q_1 \text{ and } T, A, \sigma \models Q_2, \\
T, A, \sigma \models \exists x.Q & \text{if} & \text{exists } t \in \text{ADOM}(A) \text{ such that } T, A, \sigma[x/t] \models Q,
\end{array}$$

where $\sigma[x/t]$ denotes the substitution obtained from $\sigma$ by assigning to $x$ the constant/term $t$ (if $x$ is already present in $\sigma$ its value is replaced by $t$, if not, the pair $x/t$ is added to the substitution).

The *certain answer to $Q$ over* $(T, A)$, denoted $\text{ANS}(Q, T, A)$, is the set of substitutions $\sigma$ for the free variables in $Q$ such that $Q$ holds in $(T, A)$ under $\sigma$, i.e.,

$$\text{ANS}(Q, T, A) = \{\sigma \mid T, A, \sigma \models Q\}.$$

Following the line of the proof by Calvanese et al. (2007a), but considering Theorem 2 for the basic step of evaluating an UCQ, we get:

**Theorem 3** *Computing* $\text{ANS}(Q, T, A)$ *of an ECQ $Q$ over a DL-Lite$_{\text{NU}}$ KB $(T, A)$ is PTIME-complete in the size of $T$ and $A$.*

We recall that *DL-Lite* enjoys a *rewritability* property, which states that for every UCQ $q$ and for every *DL-Lite* KB $(T, A)$,

$$\text{ANS}(q, T, A) = \text{ANS}(rew_T(q), \emptyset, A),$$

where $rew_T(q)$ is a UCQ computed by the reformulation algorithm of Calvanese et al. (2007b). Notice that, in this way, we have "compiled away" the TBox. This result can be extended to ECQs as well, i.e., for every ECQ $Q$, $\text{ANS}(Q, T, A) = \text{ANS}(rew_T(Q), \emptyset, A)$ where the query $rew_T(Q)$ is obtained from $Q$ by substituting each atom $[q]$ (where $q$ is an UCQ) by $[rew_T(q)]$ (Calvanese et al., 2007a). In our setting, we can again exploit rewritability, but only after having pre-processed the ABox (in PTIME) by propagating equalities between individual terms in $\text{ADOM}(A)$ according to functionality assertions and congruence of terms.

We say that two ABoxes $A_1$ and $A_2$ are equivalent w.r.t. TBox $T$ and predicate alphabet $\Lambda$, denoted by

$$A_1 \equiv_{T,\Lambda} A_2,$$

if for every ABox assertion $\alpha_2 \in A_2$ which is either a concept assertion $N(t)$ with $N \in \Lambda$, role assertion $P(t_1, t_2)$ with $P \in \Lambda$, or equivalence assertion $t_1 = t_2$, we have $(T, A_1) \models \alpha_2$; and vice-versa, for every ABox assertion $\alpha_1 \in A_1$, which is either a concept assertion $N(t)$ with $N \in \Lambda$, role assertion $P(t_1, t_2)$ with $P \in \Lambda$, or equivalence assertion $t_1 = t_2$, we have $(T, A_2) \models \alpha_1$. Notice that if $A_1 \equiv_{T,\Lambda} A_2$, then for every ECQ $Q$ whose concept and role names belong to $\Lambda$ we have that $\text{ANS}(Q, T, A_1) = \text{ANS}(Q, T, A_2)$. Notice also that, by applying Theorem 3 to the boolean query $[\alpha]$ corresponding to the ABox assertion $\alpha$, for each $\alpha$ in $A_1$ and $A_2$, we obtain that ABox equivalence can be checked in PTIME.

## 3. Knowledge and Action Bases

A *Knowledge and Action Base (KAB)* is a tuple $\mathcal{K} = (T, A_0, \Gamma, \Pi)$ where $T$ and $A_0$ form the *knowledge component* (or knowledge base), and $\Gamma$ and $\Pi$ form the *action component* (or action base). In practice, $\mathcal{K}$ stores the information of interest into a KB, formed by a fixed TBox $T$ and an initial ABox $A_0$, which evolves by executing actions $\Gamma$ according to the sequencing established by process $\Pi$. During the evolution new individuals can be acquired by the KB. Such individuals are witnesses of new pieces of information inserted into the KAB from the environment the KAB runs in (i.e., the external world). We represent these new objects as function terms. As the KAB evolves, the identity of individuals should be intuitively preserved and this induces the necessity of remembering equalities between terms denoting individuals discovered in the past. We describe in detail the components of the KAB.

### 3.1 TBox

$T$ is a *DL-Lite$_{\text{NU}}$* TBox, used to capture the intensional knowledge about the domain of interest. Such a TBox is fixed once and for all, and does not evolve during the execution of the KAB.

### 3.2 ABox

$A_0$ is a *DL-Lite$_{\text{NU}}$* ABox, which stores the extensional information of interest. Notice that $A_0$ is the ABox of the *initial state* of the KAB, and as the KAB evolves due to the effect of actions, the ABox, which is indeed the state of the system, evolves accordingly to store up-to-date information. Through actions we acquire new information from the external world by using calls to external services represented through functions. Given that we have no information about these services, except for their name and the parameters that are passed to them, the functions remain uninterpreted. We only assume that the result of such service calls depends only on the passed parameters. Hence, we represent the new individuals returned by service calls as function terms. The presence of function terms has an impact on the treatment of equality, since in principle we need to close equality w.r.t. congruence. While this closure generates an infinite number of logically implied equality assertions, we are going to keep such assertions implicit, computing them only when needed.

### 3.3 Actions

$\Gamma$ is a finite set of actions. An *action* $\gamma \in \Gamma$ modifies the current ABox $A$ by adding or deleting assertions, thus generating a new ABox $A'$. An action $\gamma$ has the form

$$act(\vec{x}) : \{e_1, \ldots, e_n\},$$

where $act(\vec{x})$ is the *signature* of $\gamma$ and $\{e_1, \ldots, e_n\}$ is a (finite) set of *effects* forming the *effect specification* of $\gamma$. The action signature is constituted by a name $act$ and a list $\vec{x}$ of individual *input parameters*, which need to be instantiated with actual individuals at execution time.[2] An *effect* $e_i$ has the form

$$[q_i^+] \wedge Q_i^- \rightsquigarrow A_i', \tag{1}$$

where

- $q_i^+$ is an UCQ, i.e., a positive query, which extracts the bulk data to process (obtained as the certain answers of $q_i^+$); the free variables of $q_i^+$ include the action parameters;

- $Q_i^-$ is an arbitrary ECQ, whose free variables occur all among the free variables of $q_i^+$, which refines, by using negation and quantification, the result of $q_i^+$. The query $[q_i^+] \wedge Q_i^-$ as a whole extracts individual terms to be used to form the new state of the KAB (notice that the UCQ-ECQ division is also a convenience to have readily available the positive part of the condition, which we will exploit later);

- $A_i'$ is a set of (non-ground) ABox assertions, which include as terms: constants in $A_0$, free variables of $q_i^+$, and function terms $f(\vec{x})$ having as arguments $\vec{x}$ free variables of $q_i^+$. These terms, once grounded with the values extracted from $[q_i^+] \wedge Q_i^-$, give rise to (ground) ABox assertions, which contribute to form the next state of the KAB.

More precisely, given the current ABox $A$ of $\mathcal{K}$ and a substitution $\theta$ for the input parameters of the action $\gamma$, we denote by $\gamma\theta$ the action instantiated with the *actual* parameters coming from $\theta$. By firing $\gamma\theta$ on the state $A$, we get a new state $A'$ which is computed by simultaneously applying all instantiated effects of $\gamma\theta$ as follows:

- Each effect $e_i$ in $\gamma$ of form (1) extracts from $A$ the set $\text{ANS}(([q_i^+] \wedge Q_i^-)\theta, T, A)$ of tuples of terms in $\text{ADOM}(A)$ and, for each such tuple $\sigma$, asserts a set $A_i'\theta\sigma$ of ABox assertions obtained from $A_i'\theta$ by applying the substitution $\sigma$ for the free variables of $q_i^+$. For each function term $f(\vec{x})\theta$ appearing in $A_i'\theta$, a new ground term is introduced having the form $f(\vec{x})\theta\sigma$. These terms represent new "constants" coming from the external environment the KAB is running in.

  We denote by $e_i\theta(A)$ the overall set of ABox assertions, i.e.,

$$e_i\theta(A) = \bigcup_{\sigma \in \text{ANS}(([q_i^+] \wedge Q_i^-)\theta, T, A)} A_i'\theta\sigma.$$

---

2. We disregard a specific treatment of output parameters, and assume instead that the user can freely pose queries over the KB, extracting whatever information she/he is interested in.

- Moreover, let $\text{EQ}(T, A) = \{t_1 = t_2 \mid \langle t_1, t_2 \rangle \in \text{ANS}([x_1 = x_2], T, A)\}$. Observe that, due to the semantics of queries, the terms in $\text{EQ}(T, A)$ must appear explicitly in $\text{ADOM}(A)$, that is, the possibly infinite number of equalities due to congruence do not appear in $\text{EQ}(T, A)$, though they are logically implied. Hence, the equalities in $\text{EQ}(T, A)$ are the equality assertions involving terms in $\text{ADOM}(A)$ that either appear explicitly in $A$, or are obtained by closing these under functionality and congruence of terms.

The overall *effect* of the action $\gamma$ with parameter substitution $\theta$ over $A$ is the new ABox $A' = \text{DO}(T, A, \gamma\theta)$ where

$$\text{DO}(T, A, \gamma\theta) = \text{EQ}(T, A) \cup \bigcup_{1 \leq i \leq n} e_i\theta(A).$$

Notice that the presence of function terms in action effects makes the domain of the ABoxes obtained by executing actions continuously changing and in general unbounded in size. Notice also that we do have a persistence assumption on equalities, i.e., we implicitly copy all equalities holding in the current state to the new one. This implies that, as the system evolves, we acquire new information on equalities between terms, but never lose information on equalities already acquired. Finally, we observe that in the above execution mechanism no persistence/frame assumption (except for equality) is made. In principle at every move we substitute the whole old state, i.e., ABox, with a new one. On the other hand, it should be clear that we can easily write effect specifications that *copy* big chunks of the old state into the new one. For example, $[P(x, y)] \rightsquigarrow P(x, y)$ copies the entire set of assertions involving the role $P$. In some sense, the execution mechanism adopted in this paper is very basic and does not address any of the elaboration tolerance issues typical of reasoning about actions, such as the frame problem, ramification problem or qualification problem (Reiter, 2001)[3]. This is not because we consider them irrelevant, on the contrary, they are relevant and further research on such issues is desirable. We adopt this basic mechanism simply because it is general enough to expose all difficulties we need to overcome in order to get decidability of verification in this setting.

### 3.4 Process

The process component of a KAB is a possibly nondeterministic program that uses the KAB ABoxes to store its (intermediate and final) computation results, and the actions in $\Gamma$ as atomic instructions. The ABoxes can be arbitrarily queried through the KAB TBox $T$, while they can be updated only through actions in $\Gamma$. To specify such a process component we adopt a rule-based specification.

Specifically, a *process* is a finite set $\Pi$ of condition/action rules. A *condition/action rule* $\pi \in \Pi$ is an expression of the form

$$Q \mapsto \gamma,$$

where $\gamma$ is an action in $\Gamma$ and $Q$ is an ECQ, whose free variables are exactly the parameters of $\gamma$. The rule expresses that, for each tuple $\theta$ for which condition $Q$ holds, the action $\gamma$ with actual parameters $\theta$ *can* be executed. Processes do not force the execution of actions but constrain them: the user of the process will be able to choose any action that the rules forming the process allow. Moreover, our processes inherit entirely their states from the KAB knowledge component (TBox and ABox) (see, e.g., Cohn & Hull, 2009).

---

3. But see also the work by Kowalski and Sadri (2011).

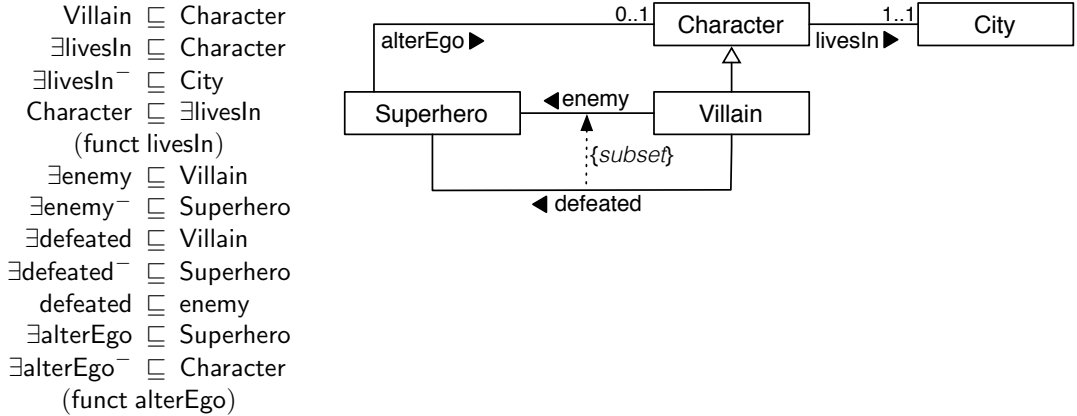

$$
\begin{aligned}
\text{Villain} &\sqsubseteq \text{Character} \\
\exists \text{livesIn} &\sqsubseteq \text{Character} \\
\exists \text{livesIn}^- &\sqsubseteq \text{City} \\
\text{Character} &\sqsubseteq \exists \text{livesIn} \\
(\text{funct livesIn}) & \\
\exists \text{enemy} &\sqsubseteq \text{Villain} \\
\exists \text{enemy}^- &\sqsubseteq \text{Superhero} \\
\exists \text{defeated} &\sqsubseteq \text{Villain} \\
\exists \text{defeated}^- &\sqsubseteq \text{Superhero} \\
\text{defeated} &\sqsubseteq \text{enemy} \\
\exists \text{alterEgo} &\sqsubseteq \text{Superhero} \\
\exists \text{alterEgo}^- &\sqsubseteq \text{Character} \\
(\text{funct alterEgo}) &
\end{aligned}
$$

Figure 1: KAB's TBox for Example 1

We observe that we adopt a basic rule-based specification here because, in spite of its simplicity, it is able to expose all the difficulties of our setting. Other choices are also possible, in particular, the process could maintain its own state besides the one of the KAB. As long as such an additional state is finite, or embeddable into the KAB itself, the results here would easily extend to such a case.

**Example 1** Let us consider a KAB $\mathcal{K} = (T, A_0, \Gamma, \Pi)$ describing a super-heroes comics world, where we have cities in which characters live. Figure 1 shows the TBox $T$ and its rendering as a UML Class Diagram. For the relationship between UML Class Diagrams and Description Logics in general and *DL-Lite* in particular, we refer to the work by Berardi, Calvanese, and De Giacomo (2005) and by Calvanese, De Giacomo, Lembo, Lenzerini, Poggi, Rodríguez-Muro, and Rosati (2009). As for the dynamics of the domain, characters can be superheroes or (super)villains, who fight each other. As in the most classic plot, superheroes help the endeavors of law enforcement fighting villains threatening the city they live in. When a villain reveals himself for perpetrating his nefarious purposes against the city's peace, he consequently becomes a declared enemy of all superheroes living in that city. Each character lives in one city at the time. A common trait of superheroes is a secret identity: a superhero is said to be the alter ego of some character, which is his identity in common life. Hence, the ABox assertion alterEgo$(s, p)$ means that the superhero $s$ is the alter ego of character $p$. Villains always try to unmask superheroes, i.e., find their secret identity, in order to exploit such a knowledge to defeat them. Notice the subtle difference here: we use the alterEgo$(s, p)$ assertion to model the fact that $s$ is the alter ego of $p$, whereas only by asserting $s = p$ we can capture the knowledge that $s$ and $p$ semantically denote the same individual. $\Gamma$ may include actions like the following ones:

$$
\begin{aligned}
\text{BecomeSH}(p, c) : \{ &[\text{Character}(p) \wedge \text{livesIn}(p, c) \wedge \exists v.\text{Villain}(v) \wedge \text{livesIn}(v, c)] \\
&\rightsquigarrow \{\text{Superhero}(\text{sh}(p)), \text{alterEgo}(\text{sh}(p), p)\}, \\
&CopyAll \}
\end{aligned}
$$

states that if there exists at least one villain living in the city $c$, a new superhero $\text{sh}(p)$ can be created, with the purpose of protecting $c$. Such a superhero has $p$ as alter ego. $CopyAll$ is a shortcut for explicitly copying all concept and role assertions to the new state (equality assertions are always implicitly copied). Action

$$\text{Unmask}(s,p): \quad \{\ [\text{alterEgo}(s,p)] \rightsquigarrow \{s = p\},$$
$$CopyAll\ \}$$

states that superhero $s$, who is the alter ego of $p$, gets unmasked by asserting the equality between $s$ and $p$ (it is now known that $s = p$). Action

$$\text{Fight}(v,s): \quad \{\ \exists p.[\text{Villain}(v) \land \text{Character}(p) \land \text{alterEgo}(s,p)] \land [s = p] \rightsquigarrow \{\text{defeated}(v,s)\},$$
$$CopyAll\ \}$$

states that when villain $v$ fights superhero $s$, he defeats $s$ if $s$ has been unmasked, i.e., it is known that $s$ is equal to his alter ego. Action

$$\text{Challenge}(v,s):$$
$$\{\ [\text{Villain}(v) \land \text{Superhero}(s) \land \exists p.\text{alterEgo}(s,p) \land \text{livesIn}(p,sc)] \land \neg[\text{defeated}(v,s)]$$
$$\rightsquigarrow \{\text{livesIn}(v,sc), \text{enemy}(v,s)\},$$
$$CopyAll\ \}$$

states that when villain $v$ challenges superhero $s$ and has not defeated him, next he lives in the same city as $s$ and is an enemy of $s$. Action

$$\text{ThreatenCity}(v,c):$$
$$\{\ [\text{Villain}(v) \land \text{Superhero}(s) \land \exists p.\text{alterEgo}(s,p) \land \text{livesIn}(p,c)]$$
$$\rightsquigarrow \{\text{enemy}(v,s) \land \text{livesIn}(v,c)\}$$
$$CopyAll\ \}$$

states that when villain $v$ threatens city $c$, then he becomes an enemy of all and only superheroes that live in $c$.

A process $\Pi$ might include the following rules:

$$[\text{Character}(p)] \land \neg[\exists s.\text{Superhero}(s) \land \text{livesIn}(s,c)] \mapsto \text{BecomeSH}(p,c),$$
$$[\text{Superhero}(s) \land \text{Character}(c)] \mapsto \text{Unmask}(s,c),$$
$$[\text{enemy}(v,s)] \land \neg[\exists v'.\text{defeated}(v',s)] \mapsto \text{Fight}(v,s),$$
$$[\text{Villain}(v) \land \text{Superhero}(s)] \mapsto \text{Challenge}(v,s),$$
$$[\text{Villain}(v) \land \text{City}(c)] \land \neg\exists v'([\text{Villain}(v') \land \text{livesIn}(v',c)] \land \neg[v = v']) \mapsto \text{ThreatenCity}(v,c).$$

For instance, the first rule states that a character can become a superhero if the city does not already have one, whereas the last one states that a villain can threaten a city, if the city does not have another villain that is (known to be) distinct from him/her.

Notice that, during the execution, reasoning on the KB is performed. For instance, consider an initial ABox

$$A_0 = \{\ \text{Superhero}(\text{batman}), \text{Villain}(\text{joker}), \text{alterEgo}(\text{batman}, \text{bruce}),$$
$$\text{livesIn}(\text{bruce}, \text{gotham}), \text{livesIn}(\text{batman}, \text{gotham}), \text{livesIn}(\text{joker}, \text{city1})\ \}.$$

In this state, bruce and batman live in the same city, and batman is the alter-ego of bruce, but it is not known whether they denote the same individual. Executing Challenge(joker, batman) in $A_0$, which is indeed allowed by the process $\Pi$, generates a new ABox with added assertions enemy(joker, batman), livesIn(joker, gotham), and gotham = city1 is implied by the functionality on livesIn.  □

## 4. KAB Transition System

The semantics of KABs is given in terms of possibly infinite transition systems that represent the possible evolutions of the KAB over time, as actions are executed according to the process. Notice that such transition systems must be equipped with semantically rich states, since a full KB is associated to them. Formally we define the kind of transition system we need as follows: A *transition system* $\Upsilon$ is a tuple of the form $(\mathbb{U}, T, \Sigma, s_0, abox, \Rightarrow)$, where:

- $\mathbb{U}$ is a countably infinite set of terms denoting individuals, called *universe*;
- $T$ is a TBox;
- $\Sigma$ is a set of states;
- $s_0 \in \Sigma$ is the initial state;
- $abox$ is a function that, given a state $s \in \Sigma$ returns an ABox associated to $s$ which has as individuals terms of $\mathbb{U}$, and which conforms to $T$;
- $\Rightarrow \subseteq \Sigma \times \Sigma$ is a transition relation between pairs of states.

For convenience, we introduce the active domain of the whole transition system, defined as $\text{ADOM}(\Upsilon) = \bigcup_{s \in \Sigma} \text{ADOM}(abox(s))$. Also we introduce the (predicate) alphabet $\text{ALPH}(\Upsilon)$ of $\Upsilon$ as the set of concepts and roles occurring in $T$ or in the co-domain of $abox$.

The KAB generates a transition system of this form during its execution. Formally, given a KAB $\mathcal{K} = (T, A_0, \Gamma, \Pi)$, we define its *(generated) transition system* $\Upsilon_\mathcal{K} = (\mathbb{U}, T, \Sigma, s_0, abox, \Rightarrow)$ as follows:

- $\mathbb{U}$ is formed by all constants and all function terms inductively formed starting from $\text{ADOM}(A_0)$ by applying the functions occurring in the actions in $\Gamma$;
- $T$ is the TBox of the KAB;
- $abox$ is the identity function (i.e., each state is simply an ABox);
- $s_0 = A_0$ is the initial state;
- $\Sigma$ and $\Rightarrow$ are defined by mutual induction as the smallest sets satisfying the following property: if $s \in \Sigma$, then for each rule $Q \mapsto \gamma$, evaluate $Q$ and, for each tuple $\theta$ returned, if $\text{DO}(T, abox(s), \gamma\theta)$ is consistent w.r.t. $T$, then $s \Rightarrow s'$ where $s' = \text{DO}(T, abox(s), \gamma\theta)$.

Notice that the alphabet $\text{ALPH}(\Upsilon_\mathcal{K})$ of $\Upsilon_\mathcal{K}$ is simply formed by the set $\text{ALPH}(\mathcal{K})$ of concepts and roles that occur in $\mathcal{K}$.

The KAB transition system $\Upsilon_\mathcal{K}$ is an infinite tree with infinitely many different ABoxes in its nodes, in general. In fact, to get a transition system that is infinite, it is enough to perform indefinitely a simple action that adds new terms at each step, e.g., an action of the form

$$\gamma() : \{\ [C(x)] \rightsquigarrow \{C(f(x))\},\ \ CopyAll\ \}.$$

Hence the classical results on model checking (Clarke et al., 1999), which are developed for finite transition systems, cannot be applied directly for verifying KABs.

## 5. Verification Formalism

To specify dynamic properties over KABs, we use a first-order variant of $\mu$-calculus (Stirling, 2001; Park, 1976). $\mu$-calculus is virtually the most powerful temporal logic used for model checking of finite-state transition systems, and is able to express both linear time logics such as LTL and PSL, and branching time logics such as CTL and CTL* (Clarke et al., 1999). The main characteristic of $\mu$-calculus is its ability of expressing directly least and greatest fixpoints of (predicate-transformer) operators formed using formulae relating the current state to the next one. By using such fixpoint constructs one can easily express sophisticated properties defined by induction or co-induction. This is the reason why virtually all logics used in verification can be considered as fragments of $\mu$-calculus. Technically, $\mu$-calculus separates local properties, asserted on the current state or on states that are immediate successors of the current one, from properties talking about states that are arbitrarily far away from the current one (Stirling, 2001). The latter are expressed through the use of fixpoints.

In this work, we use a first-order variant of $\mu$-calculus, where we allow local properties to be expressed as ECQs, and at the same time we allow for arbitrary first-order quantification across states. Given the nature of ECQs used for formulating local properties, first-order quantification ranges over terms denoting individuals. Formally, we introduce the logic $\mu\mathcal{L}_A$ defined as follows:

$$\Phi \longrightarrow Q \mid \neg\Phi \mid \Phi_1 \wedge \Phi_2 \mid \exists x.\Phi \mid \langle-\rangle\Phi \mid Z \mid \mu Z.\Phi,$$

where $Q$ is a possibly open ECQ and $Z$ is a second order predicate variable (of arity 0). We make use of the following abbreviations: $\forall x.\Phi = \neg(\exists x.\neg\Phi)$, $\Phi_1 \vee \Phi_2 = \neg(\neg\Phi_1 \wedge \neg\Phi_2)$, $[-]\Phi = \neg\langle-\rangle\neg\Phi$, and $\nu Z.\Phi = \neg\mu Z.\neg\Phi[Z/\neg Z]$. The formulae $\mu Z.\Phi$ and $\nu Z.\Phi$ respectively denote the least and greatest fixpoint of the formula $\Phi$ (seen as the predicate transformer $\lambda Z.\Phi$). As usual in $\mu$-calculus, formulae of the form $\mu Z.\Phi$ (and $\nu Z.\Phi$) must obey to the *syntactic monotonicity* of $\Phi$ w.r.t. $Z$, which states that every occurrence of the variable $Z$ in $\Phi$ must be within the scope of an even number of negation symbols. This ensures that the least fixpoint $\mu Z.\Phi$ (as well as the greatest fixpoint $\nu Z.\Phi$) always exists.

The semantics of $\mu\mathcal{L}_A$ formulae is defined over possibly infinite transition systems of the form $\langle \mathbb{U}, T, \Sigma, s_0, abox, \Rightarrow \rangle$ seen above. Since $\mu\mathcal{L}_A$ also contains formulae with both individual and predicate free variables, given a transition system $\Upsilon$, we introduce an *individual variable valuation* $v$, i.e., a mapping from individual variables $x$ to $\mathbb{U}$, and a *predicate variable valuation* $V$, i.e., a mapping from the predicate variables $Z$ to subsets of $\Sigma$. With these three notions in place, we assign meaning to formulae by associating to $\Upsilon$, $v$, and $V$ an *extension function* $(\cdot)^{\Upsilon}_{v,V}$, which maps formulae to subsets of $\Sigma$. Formally, the extension function $(\cdot)^{\Upsilon}_{v,V}$ is defined inductively as follows:

$$\begin{aligned}
(Q)^{\Upsilon}_{v,V} &= \{s \in \Sigma \mid \textsc{ans}(Qv, T, abox(s)) = \mathit{true}\}, \\
(\neg\Phi)^{\Upsilon}_{v,V} &= \Sigma \setminus (\Phi)^{\Upsilon}_{v,V}, \\
(\Phi_1 \wedge \Phi_2)^{\Upsilon}_{v,V} &= (\Phi_1)^{\Upsilon}_{v,V} \cap (\Phi_2)^{\Upsilon}_{v,V}, \\
(\exists x.\Phi)^{\Upsilon}_{v,V} &= \{s \in \Sigma \mid \exists t.t \in \textsc{adom}(abox(s)) \text{ and } s \in (\Phi)^{\Upsilon}_{v[x/t],V}\}, \\
(\langle-\rangle\Phi)^{\Upsilon}_{v,V} &= \{s \in \Sigma \mid \exists s'.s \Rightarrow s' \text{ and } s' \in (\Phi)^{\Upsilon}_{v,V}\}, \\
(Z)^{\Upsilon}_{v,V} &= V(Z), \\
(\mu Z.\Phi)^{\Upsilon}_{v,V} &= \bigcap\{\mathcal{E} \subseteq \Sigma \mid (\Phi)^{\Upsilon}_{v,V[Z/\mathcal{E}]} \subseteq \mathcal{E}\}.
\end{aligned}$$

Here $Qv$ stands for the (boolean) ECQ obtained from $Q$ by substituting its free variables according to $v$. Intuitively, $(\cdot)^{\Upsilon}_{v,V}$ assigns to such constructs the following meaning:

- The boolean connectives have the expected meaning.

- The quantification of individuals is done over the terms of the "current" ABox. Notice that such terms can be referred in later states where the associated ABox does not include such terms anymore.

- The extension of $\langle - \rangle \Phi$ consists of the states $s$ such that, for *some* state $s'$ with transition $s \Rightarrow s'$, the formula $\Phi$ holds in $s'$ under valuation $v$.

- The extension of $[-]\Phi$ consists of the states $s$ such that, for *all* states $s'$ with transition $s \Rightarrow s'$, the formula $\Phi$ holds in $s'$ under valuation $v$.

- The extension of $\mu Z.\Phi$ is the *smallest subset* $\mathcal{E}_\mu$ of $\Sigma$ such that, when assigning to $Z$ the extension $\mathcal{E}_\mu$, the resulting extension of $\Phi$ (under valuation $v$) is contained in $\mathcal{E}_\mu$. That is, the extension of $\mu Z.\Phi$ is the *least fixpoint* of the operator $(\Phi)^\Upsilon_{v,V[Z/\mathcal{E}]}$, where $V[Z/\mathcal{E}]$ denotes the predicate valuation obtained from $V$ by forcing the valuation of $Z$ to be $\mathcal{E}$.

- Similarly, the extension of $\nu Z.\Phi$ is the *greatest subset* $\mathcal{E}_\nu$ of $\Sigma$ such that, when assigning to $Z$ the extension $\mathcal{E}_\nu$, the resulting extension of $\Phi$ contains $\mathcal{E}_\nu$. That is, the extension of $\nu Z.\Phi$ is the *greatest fixpoint* of the operator $(\Phi)^\Upsilon_{v,V[Z/\mathcal{E}]}$. Formally, $(\nu Z.\Phi)^\Upsilon_{v,V} = \bigcup \{\mathcal{E} \subseteq \Sigma \mid \mathcal{E} \subseteq (\Phi)^\Upsilon_{v,V[Z/\mathcal{E}]}\}$.

When $\Phi$ is a closed formula, $(\Phi)^\Upsilon_{v,V}$ does not depend on $v$ or $V$, and we denote the extension of $\Phi$ simply by $(\Phi)^\Upsilon$. A closed formula $\Phi$ holds in a state $s \in \Sigma$ if $s \in (\Phi)^\Upsilon$. In this case, we write $\Upsilon, s \models \Phi$. A closed formula $\Phi$ holds in $\Upsilon$, denoted by $\Upsilon \models \Phi$, if $\Upsilon, s_0 \models \Phi$. We call *model checking* the problem of verifying whether $\Upsilon \models \Phi$ holds.

The next example shows some simple temporal properties that can be expressed in $\mu\mathcal{L}_A$.

**Example 2** Considering the KAB of Example 1, we can easily express temporal properties as the following ones.

- From now on all current superheroes that live in Gotham will live in Gotham forever (a form of safety):

$$\forall x.[\mathsf{Superhero}(x) \wedge \mathsf{livesIn}(x, \mathsf{gotham})] \supset \nu Z.([\mathsf{livesIn}(x, \mathsf{gotham})] \wedge [-]Z).$$

- Eventually all current superheroes will be unmasked (a form of liveness):

$$\forall x.[\mathsf{Superhero}(x)] \supset \mu Z.([\mathsf{alterEgo}(x,x)] \vee [-]Z).$$

- There exists a possible future situation where all current superheroes will be unmasked (another form of liveness):

$$\forall x.[\mathsf{Superhero}(x)] \supset \mu Z.([\mathsf{alterEgo}(x,x)] \vee \langle - \rangle Z).$$

- Along every future, it is always true, for every superhero, that there exists an evolution that eventually leads to unmask him (a form of liveness that holds in every moment):

$$\nu Y.(\forall x.[\mathsf{Superhero}(x)] \supset \mu Z.([\mathsf{alterEgo}(x,x)] \vee \langle - \rangle Z)) \wedge [-]Y. \qquad \square$$

Consider two transition systems sharing the same universe and the same predicate alphabet. We say that they are *behaviorally equivalent* if they satisfy exactly the same $\mu\mathcal{L}_A$ formulas. To formally capture such an equivalence, we make use of the notion of *bisimulation* (Milner, 1971), suitably extended to deal with query answering over KBs.

Given two transition systems $\Upsilon_1 = \langle \mathbb{U}, T, \Sigma_1, s_{01}, abox_1, \Rightarrow_1 \rangle$ and $\Upsilon_2 = \langle \mathbb{U}, T, \Sigma_2, s_{02}, abox_2, \Rightarrow_2 \rangle$ sharing the same universe $\mathbb{U}$, the same TBox $T$, and such that $\text{ALPH}(\Upsilon_1) = \text{ALPH}(\Upsilon_2) = \Lambda$, a *bisimulation* between $\Upsilon_1$ and $\Upsilon_2$ is a relation $\mathcal{B} \subseteq \Sigma_1 \times \Sigma_2$ such that $(s_1, s_2) \in \mathcal{B}$ implies that:

1. $abox(s_1) \equiv_{T,\Lambda} abox(s_2)$;
2. if $s_1 \Rightarrow_1 s'_1$, then there exists $s'_2$ such that $s_2 \Rightarrow_2 s'_2$ and $(s'_1, s'_2) \in \mathcal{B}$;
3. if $s_2 \Rightarrow_2 s'_2$, then there exists $s'_1$ such that $s_1 \Rightarrow_1 s'_1$ and $(s'_1, s'_2) \in \mathcal{B}$.

We say that two states $s_1$ and $s_2$ are *bisimilar* if there exists a bisimulation $\mathcal{B}$ such that $(s_1, s_2) \in \mathcal{B}$. Two transition systems $\Upsilon_1$ with initial state $s_{01}$ and $\Upsilon_2$ with initial state $s_{02}$ are *bisimilar* if $(s_{01}, s_{02}) \in \mathcal{B}$. The following theorem states that the formula evaluation in $\mu\mathcal{L}_A$ is indeed invariant w.r.t. bisimulation, so we can equivalently check any bisimilar transition systems.

**Theorem 4** *Let $\Upsilon_1$ and $\Upsilon_2$ be two transition systems that share the same universe, the same TBox, and the same predicate alphabet, and that are bisimilar. Then, for two states $s_1$ of $\Upsilon_1$ and $s_2$ of $\Upsilon_2$ (including the initial ones) that are bisimilar, and for all closed $\mu\mathcal{L}_A$ formulas $\Phi$, we have that*

$$s_1 \in (\Phi)^{\Upsilon_1} \quad \textit{iff} \quad s_2 \in (\Phi)^{\Upsilon_2}.$$

*Proof.* The proof is analogous to the standard proof of bisimulation invariance of $\mu$-calculus (Stirling, 2001), though taking into account our bisimulation, which guarantees that ECQs are evaluated identically over bisimilar states. Notice that the assumption that the two transition systems share the same universe and the same predicate alphabet makes it easy to compare the answers to queries. □

Making use of such a notion of bisimulation, we can, for example, redefine the transition system generated by a KAB $\mathcal{K} = (T, A_0, \Gamma, \Pi)$ while maintaining bisimilarity, by modifying the definition of $\Upsilon_\mathcal{K} = \langle \mathbb{U}, T, \Sigma, s_0, abox, \Rightarrow \rangle$ given in Section 4 as follows.

(i) We modify DO() so that no function term $t'$ is introduced in the generated ABox $A'$ if in the current ABox[4] $A$ there is already a term $t$ such that $(T, A) \models t = t'$.

(ii) If the ABox $A' = \text{DO}(T, abox(s), \gamma\theta)$ obtained from the current state $s$ is logically equivalent to the ABox $abox(s'')$, for some already generate state $s''$, we do not generate a new state, but simply add $s \Rightarrow s''$ to $\Upsilon_\mathcal{K}$.

## 6. Verification of KABs

It is immediate to see that verification of KABs is undecidable in general as it is easy to represent Turing machines using a KAB. Actually we can do so using only a fragment of the capabilities of KABs, as shown in the next lemma.

**Lemma 5** *Checking formulas of the form $\mu Z.(N(a) \vee \langle - \rangle Z)$, where $N$ is an atomic concept and $a$ is an individual occurring in $A_0$, is undecidable already for a KAB $\mathcal{K} = (T, A_0, \Gamma, \Pi)$ where:*

---

4. Note that all terms that are present in the current ABox are preserved in the new ABox, together with equalities between terms.

$$
\begin{aligned}
\text{[First}(c)] &\rightsquigarrow \{\text{First}(c)\} \\
[\text{cell}(c, \#) \wedge \text{value}(c, x)] &\rightsquigarrow \{\text{value}(c, x)\} \\
[\text{cell}(c, \mathsf{a}_q) \wedge \text{value}(c, \mathsf{a}_v)] &\rightsquigarrow \{\text{value}(c, \mathsf{a}_{v'})\} \\
[\text{cell}(c, \mathsf{a}_q) \wedge \text{value}(c, \mathsf{a}_v) \wedge \text{next}(c, c_r)] &\rightsquigarrow \{\text{cell}(c_r, \mathsf{a}_{q'})\} \\
[\text{cell}(c, \mathsf{a}_q) \wedge \text{value}(c, \mathsf{a}_v) \wedge \text{Last}(c)] &\rightsquigarrow \{\text{cell}(\mathsf{n}(c), \mathsf{a}_{q'}), \text{next}(c, \mathsf{n}(c)), \text{Last}(\mathsf{n}(c))\} \\
[\text{cell}(c, \#) \wedge \text{Last}(c)] &\rightsquigarrow \{\text{Last}(c)\} \\
[\text{cell}(c, \#) \wedge \text{First}(c)] &\rightsquigarrow \{\text{cell}(c, \#)\} \\
[\text{cell}(c, \#) \wedge \text{next}(c, c_r)] &\rightsquigarrow \{\text{cell}(c_r, \#)\} \\
[\text{cell}(c, \mathsf{a}_{q_f})] &\rightsquigarrow \{\text{Stop}(0)\}
\end{aligned}
$$

Figure 2: Effects of the action used to encode a transition $\delta(q, v, q', v', R)$ of a Turing Machine

- $T$ is the empty TBox,
- the actions in $\Gamma$ make no use of negation nor equality,
- $\Pi$ is the trivial process that poses no restriction on executability of actions.

*Proof.* Given a Turing machine $\mathcal{M} = \langle Q, \Sigma, q_0, \delta, q_f, \_\rangle$, we show how to construct a corresponding KAB $\mathcal{K}_\mathcal{M} = (\emptyset, A_0, \Gamma, \Pi)$ that mimics the behavior of $\mathcal{M}$. Specifically, we encode the halting problem for $\mathcal{M}$ as a verification problem over $\mathcal{K}_\mathcal{M}$. Roughly speaking, $\mathcal{K}_\mathcal{M}$ maintains the tape and state information in the (current) ABox, and encodes the transitions of $\mathcal{M}$ as actions. Our construction makes use of a tape that initially contains a unique cell, represented by the constant 0, and is extended on-the-fly as needed: cells to the right of 0 are represented by function terms of the form $\mathsf{n}(\mathsf{n}(\cdots(0)\cdots))$, while cells to the left of 0 are represented by function terms of the form $\mathsf{p}(\mathsf{p}(\cdots(0)\cdots))$. Then, we make use of one constant $\mathsf{a}_q$ for each state $q \in Q$, of one constant $\mathsf{a}_v$ for each tape symbol value $v \in \Sigma$, of a special constant $\#$, and of the following concepts and roles:

- $\text{cell}(c, h)$ models a cell of the tape, where $c$ is a cell identifier, and $h$ corresponds to the current state of $\mathcal{M}$, if the head of $\mathcal{M}$ currently points to $c$, or to $\#$ if the head does not currently point to $c$;
- $\text{next}(c_l, c_r)$ models the relative position of cells, stating that $c_r$ is the cell immediately following $c_l$;
- $\text{value}(c, v)$ models that cell $c$ currently contains value $v$, with $v \in \Sigma$;
- $\text{First}(c)$ and $\text{Last}(c)$ respectively denote the current first cell and last cell of the portion of tape explored so far.
- $\text{Stop}(c)$ is used to detect when $\mathcal{M}$ halts.

The initial state of $\mathcal{K}_\mathcal{M}$ contains a unique cell and is defined as

$$A_0 = \{\ \text{cell}(0, \mathsf{a}_{q_0}),\ \text{value}(0, \mathsf{a}_\_),\ \text{First}(0),\ \text{Last}(0)\ \}.$$

As for the action component, $\Gamma$ contains an action with no parameters for each transition in $\delta$, while the process $\Pi$ poses no restriction on executability of actions, i.e., it contains a rule $\text{true} \mapsto \gamma()$ for each such action $\gamma$.

We now provide the specification of actions, detailing the case of a right shift transition $\delta(q, v, q', v', R)$. The corresponding action specification consists of the set of effects shown in Figure 2. The first effect maintains the first position of the tape unaltered. The second and third effects deal with the cell values. They remain the same except for the current cell, that is updated according to the transition. The next three effects deal with the right shift and the Turing Machine state. If the current cell has a next cell and therefore is not the last one, then the head is moved to the next cell and the state change of $\mathcal{M}$ is recorded there. In this case the last cell remains the same. If instead the current cell is the last one, before moving the head the tape must be properly extended. The function n/1 is used to create the identifier of this new successor cell, starting from the identifier of the current one. Furthermore, since the transition corresponds to a right shift of one cell, the first cell and all the cells immediately following a cell marked # will be marked # in the next state. Finally, the last effect is used to identify the case in which $\mathcal{M}$ has reached a final state. This is marked by inserting into the new state the special assertion Stop(0).

The construction for a left shift transition is done symmetrically, using the function p/1 to create a new predecessor cell. By construction, $\mathcal{K}_\mathcal{M}$ satisfies the conditions of the theorem. Observe that, in the transition system $\Upsilon_{\mathcal{K}_\mathcal{M}}$ generated by $\mathcal{K}_\mathcal{M}$, every action corresponding to every transition of $\mathcal{M}$ can be executed in each ABox/state $s$ of $\Upsilon_{\mathcal{K}_\mathcal{M}}$, and since $T$ is empty, it will actually generate a successor state of $s$. However, in each state, only the (unique) action that corresponds to the actually executed transition of $\mathcal{M}$ will generate a successor state containing an ABox assertion of the form cell$(c, \mathsf{a}_q)$, for some state $q$ of $\mathcal{M}$. Therefore, only those ABoxes/states properly corresponding to configurations of $\mathcal{M}$ could eventually lead to an ABox/state in $\Upsilon_{\mathcal{K}_\mathcal{M}}$ where Stop(0) holds. And the latter will happen if and only if $\mathcal{M}$ halts. More precisely, one can show by induction on the length respectively of a halting computation of $\mathcal{M}$ and of the shortest path from the initial state of $\Upsilon_{\mathcal{K}_\mathcal{M}}$ to a state where Stop(0) holds, that $\mathcal{M}$ halts if and only if $\mathcal{K}_\mathcal{M} \models \mu Z.([\mathsf{Stop}(0)] \vee \langle - \rangle Z)$, which concludes the proof. □

From the previous lemma, which shows undecidability already in a special case, we immediately obtain the following result.

**Theorem 6** *Verification of $\mu\mathcal{L}_A$ formulae over KABs is undecidable.*

We observe that Lemma 5 uses a KB that is constituted only by an ABox containing concept and role assertions, and makes use only of conjunctive queries in defining actions effects. Moreover, the formula that we check makes no use of quantification at all, and can simply be seen as a propositional CTL formula of the form $EFp$, expressing that proposition $p$ eventually holds along one path.

## 7. Verification of Weakly Acyclic KABs

In spite of Theorem 6, next we introduce a notable class of KABs for which verification of arbitrary $\mu\mathcal{L}_A$ properties is decidable. To do so, we rely on a syntactic restriction that resembles the notion of *weak acyclicity* in data exchange (Fagin et al., 2005)[5], and that guarantees boundedness of ABoxes generated by the execution of the KAB and, in turn, decidability of verification.

Now we are ready to introduce the notion of weak acyclicity in our context. We introduce the edge-labeled directed *dependency graph* of a KAB $\mathcal{K} = (T, A_0, \Gamma, \Pi)$, defined as follows. *Nodes*, called *positions*, are obtained from the TBox T: there is a node for every concept name $N$ in $T$, and two nodes for every role name $P$ in $T$, corresponding to the domain and to the range of $P$. *Edges* are drawn by considering every effect specification $[q^+] \wedge Q^- \rightsquigarrow A'$ of each action contained in $\Gamma$, tracing how values are copied or contribute to generate new values as the system progresses. In particular, let $p$ be a position corresponding to a concept/role component in the rewriting $rew_T(q^+)$ of $q^+$ with variable $x$. For every position $p'$ in $A'$ with the same variable $x$, we include a *normal edge* $p \rightarrow p'$. For every position $p''$ in $A'$ with a function term $f(\vec{t})$ such that $x \in \vec{t}$, we include a *special edge* $p \xrightarrow{*} p''$. We say that $\mathcal{K}$ is *weakly-acyclic* if its dependency graph has no cycle going through a special edge.

---

5. We use the original definition of weak acyclicity. However, our results can be applied also to other variants of weak acyclicity (see discussion in Section 9).

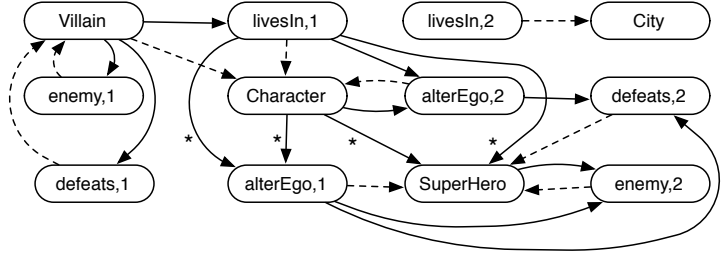

Figure 3: Weakly acyclic dependency graph for Example 1.

**Example 3** The KAB of Example 1 is weakly acyclic. Its dependency graph, shown in Figure 3, does not contain any cycle going through special edges. For readability, self-loops are not shown in the Figure (but are present for all nodes), and dashed edges are used to compactly represent the contributions given by the rewriting of the queries. E.g., the dashed edge form Villain to Character denotes that for every outgoing edge from Character, there exists an outgoing edge from Villain with the same type and target. Hence, w.r.t. weak acyclicity dashed edges can be simply replaced by normal edges. □

We are now ready to state the main result of this work, which we are going to prove in the remainder of this section.

**Theorem 7** *Verification of $\mu\mathcal{L}_A$ properties for a weakly acyclic KAB is decidable in* EXPTIME *in the size of the KAB.*

We observe that the restriction imposed by weak acyclicity (or variants) is not too severe, and in many real cases KABs are indeed weakly acyclic or can be transformed into weakly acyclic ones at cost of redesign. Indeed, weakly acyclic KABs cannot indefinitely generate new values from the old ones, which then depend on a chain of unboundedly many previous values. In other words, current values depend only on a bounded number of old values. While unbounded systems exist in theory, e.g., Turing machines, higher level processes, as those in business process management or service-oriented modeling, typically require such a boundedness in practice. How to systematically transform systems into weakly acyclic ones remains an open issue.

In the remainder of this section we present the proof of Theorem 7. We do so in several steps:

1. **Normalized KAB.** First we introduce a normalized form $\hat{\mathcal{K}}$ of the KAB $\mathcal{K}$, which isolates the contribution of equalities and of the TBox in actions effects of the KAB. An important point is that normalizing the KAB preserves weak acyclicity.

2. **Normalized DO().** Then, we introduce a normalized version $\textsc{do}_{\textsc{norm}}()$ of DO(), which avoids to consider equalities in generating the bulk set of tuples to be used in the effects to generate the next ABox. The transition system $\Upsilon_{\hat{\mathcal{K}},\textsc{norm}}$ generated through this normalized version $\textsc{do}_{\textsc{norm}}()$ of DO() by the normalized KAB $\hat{\mathcal{K}}$ is bisimilar to the transition system $\Upsilon_\mathcal{K}$ generated through DO() by $\mathcal{K}$. Hence the two transition systems satisfy the same $\mu\mathcal{L}_A$ formulae.

3. **Positive dominant.** The next step is to introduce what we call the *positive dominant* $\mathcal{K}^{++}$ of the normalized KAB $\hat{\mathcal{K}}$. This is obtained from $\hat{\mathcal{K}}$ essentially by dropping equalities, negations, and TBox. However $\mathcal{K}^{++}$ contains enough information in the positive part so that, when we drop all of these features, the active domain of the transition system $\Upsilon_{\mathcal{K}^{++}}$ generated by $\mathcal{K}^{++}$ "overestimates" the active domain of the transition system $\Upsilon_{\hat{\mathcal{K}},\textsc{norm}}$ generated by the normalized KAB $\hat{\mathcal{K}}$. Moreover, if the normalized (and hence the original) KAB is weakly acyclic, so is its positive dominant. Finally if the positive dominant is weakly acyclic then the size of the active domain of its transition system $\Upsilon_{\mathcal{K}^{++}}$ is polynomially bounded by the size of its initial ABox, and hence so is the size of the active domain of $\Upsilon_{\hat{\mathcal{K}},\textsc{norm}}$. This implies that the size of $\Upsilon_{\hat{\mathcal{K}},\textsc{norm}}$ is finite and at most exponential in the size of its initial ABox.

4. **Putting it all together.** Tying these results together, we get the claim.

In the following, we detail each of these steps.

### 7.1 Normalized KAB

Given a KAB $\mathcal{K} = (T, A_0, \Gamma, \Pi)$, we build a KAB $\hat{\mathcal{K}} = (T, \hat{A}_0, \hat{\Gamma}, \Pi)$, called the *normalized form of* $\mathcal{K}$, by applying a sequence of transformations that preserve the semantics of $\mathcal{K}$ while producing a KAB of a format that is easier to study.

1. We view each ABox $A$ as partitioned into a part collecting all concept and role assertions, and a part collecting all equality assertions. We denote with $A^{\neq Q}$ the former and with $\textsc{eq}(T, A)$ the latter, after having closed it w.r.t. (the functionality assertions in) the TBox $T$. Notice that such a closure can be computed in polynomial time in the size of $A$ and $T$.

2. In $\hat{\mathcal{K}}$ all individuals appearing in equality assertions in an ABox also occur in special concept assertions of the form $Dummy(t)$, where the concept $Dummy$ is unrelated to the other concepts and roles in the KAB. We do so by:
   - adding concept assertions $Dummy(t)$ for each $t$ in an equality assertion in $A_0$ that does not appear elsewhere;
   - adding to the right-hand part of each action effect $e_i$ a concept assertion $Dummy(t)$ for each $t$ in an equality assertion in the right-hand part of $e_i$;
   - adding to each action an effect specification of the form

   $$[Dummy(x)] \rightsquigarrow \{Dummy(x)\}.$$

   Notice that, as the result of this transformation, we get ABoxes containing the additional concept $Dummy$, which however is never queried by actions effects and by the rules forming the process. The impact of the transformation is simply that now the ADOM($A$) of the ABoxes $A$ in the KAB transition system can be readily identified as the set of terms occurring in concept and role assertion only (without looking at equality assertions).

Given an ABox $A$, we denote by $\hat{A}$ the result of the two above transformations, which respectively add to $A$ the closure of equalities and the extension of $Dummy$.

3. We manipulate each resulting effect specification

$$[q^+] \wedge Q^- \rightsquigarrow \hat{A}'$$

as follows:

3.1. We replace $[q^+] \wedge Q^-$ by $[rew_T(q^+)] \wedge rew_T(Q^-)$ (Calvanese et al., 2007a), exploiting the results by Calvanese et al. (2007b) and by Artale et al. (2009), which guarantee that, for every ECQ $Q$ and every ABox $A$ where equalities are closed under functionality and congruence, we have that

$$\text{ANS}(Q, T, A) = \text{ANS}(rew_T(Q), \emptyset, A).$$

3.2. We replace each effect specification $[rew_T(q^+)] \wedge rew_T(Q^-) \rightsquigarrow \hat{A}'$, resulting from Step 3.1, by a set of effect specifications $[q_i^+] \wedge rew_T(Q^-) \rightsquigarrow \hat{A}'$, one for each CQ $q_i$ in the UCQ $rew_T(q^+)$.

3.3. For each effect specification $[q_i^+] \wedge rew_T(Q^-) \rightsquigarrow \hat{A}'$, we re-express $q_i^+$ so as to make equalities used to join terms explicit and so as to remove constants from $q_i^+$. Specifically, we replace the effect specification by

$$[q_i^{++}] \wedge q^= \wedge rew_T(Q^-) \rightsquigarrow \hat{A}',$$

where:

- $q_i^{++}$ is the CQ without repeated variables obtained from $q_i^+$ by *(i)* replacing for each variable $x$ occurring in $q_i^+$, the $j$-th occurrence of $x$ except for the first one, by $x^{[j]}$; and *(ii)* replacing each constant $c$ with a new variable $x_c$;
- $q^= = \bigwedge [x = x^{[j]}] \wedge \bigwedge [x_c = c]$ where *(i)* the first conjunction contains one equality $[x = x^{[j]}]$ for each variable $x$ in $q_i^+$ and for each variable $x^{[j]}$ introduced in the step above, and *(ii)* the second conjunction contains one equality for each constant $c$ in $q_i^+$.

To clarify the latter consider the following example:

**Example 4** Given a query

$$[q_i^+] \doteq [N(x) \wedge P_1(x, y) \wedge P_2(c, x)],$$

Step 3.3 above replaces it by $[q_i^{++}] \wedge q^=$, where

$$q_i^{++} \doteq N(x) \wedge P_1(x^{[2]}, y) \wedge P_2(x_c, x^{[3]}), \qquad q^= \doteq [x = x^{[2]}] \wedge [x = x^{[3]}] \wedge [x_c = c]. \quad \square$$

As for the correctness of Step 3.3, it is immediate to notice that $[q_i^+]$ is equivalent to $[q_i^{++} \wedge \bigwedge(x = x^{[j]}) \wedge \bigwedge(x_c = c)]$. The equivalence between the latter and $[q_i^{++}] \wedge q^=$ is a consequence of the construction by Artale et al. (2009), which shows that query entailment in the presence of equalities can be reduced to query evaluation by saturating equalities w.r.t. transitivity, reflexivity, symmetry, and functionality.

Given an action $\gamma$, we denote by $\hat\gamma$ the action normalized as above.

Since all transformations preserve logical equivalence (as long as we do not query $Dummy$), we have

**Lemma 8** $\text{DO}(T, A, \gamma\theta) \equiv_{T,\text{ALPH}(\mathcal{K})} \text{DO}(T, \hat{A}, \hat\gamma\theta)$.

Also the normalization of a KAB preserves weak acyclicity, which is a crucial consideration for later results.

**Lemma 9** *If $\mathcal{K}$ is weakly acyclic, then also $\hat{\mathcal{K}}$ is weakly acyclic.*

*Proof.* Consider each effect specification $[q^+] \wedge Q^- \rightsquigarrow A'$ belonging to an action in $\mathcal{K}$. The contribution of this effect specification to the dependency graph $\mathcal{G}$ of $\mathcal{K}$ is limited to each CQ $q_i$ in the UCQ $rew_T(q^+)$, and to the set of concept and role assertions of $A'$. We observe that each such $q_i$ corresponds to a query $q_i^{++}$ in $\hat{\mathcal{K}}$ in which each variable of $q_i$ occurs exactly once. For every free variable $x$ of $q_i$ that also appears in $A'$, and for every occurrence of $x$ in $q_i$ itself, an edge is included in $\mathcal{G}$. In the dependency graph $\hat{\mathcal{G}}$ of $\hat{\mathcal{K}}$, only one of such edges appears, corresponding to the single occurrence of the variable $x$ in $q_i^{++}$.

Notice that $Dummy$ can be omitted from the dependency graph of $\hat{\mathcal{G}}$ since, by definition of $\hat{\mathcal{K}}$, $Dummy$ does not occur in the left-hand side of effects except for the trivial effect $[Dummy(x)] \rightsquigarrow \{Dummy(x)\}$. This is not true for $\mathcal{K}$, where $Dummy$ is needed. Therefore, $\hat{\mathcal{G}}$ is indeed a subgraph of $\mathcal{G}$, and hence weak acyclicity of $\mathcal{G}$ implies weak acyclicity of $\hat{\mathcal{G}}$. □

### 7.2 Normalized DO()

Next we give a simplified version of DO(), which we call $\text{DO}_{\text{NORM}}()$. We start by observing that we can reformulate the definition of DO() given in Section 3. For that, we first need to define a suitable notion of *join* of two queries. Let $q_1$ and $q_2$ be two ECQs, which may have free variables in common, and let $A_1$ and $A_2$ be two ABoxes. Then we define $\text{ANS}(q_1, \emptyset, A_1) \bowtie \text{ANS}(q_2, \emptyset, A_2)$ as the set of substitutions $\sigma$ over the free variables in $q_1$ and $q_2$ such that $q_i$ holds in $\emptyset, A_i$ under $\sigma$, i.e., $\emptyset, A_i, \sigma \models q_i$, for $i \in \{1, 2\}$. Then, given an action $\hat\gamma$ with parameters substitution $\theta$ and an ABox $\hat{A}$, we have

$$\text{DO}(T, \hat{A}, \hat\gamma\theta) = \bigcup_{e \text{ in } \hat\gamma} \text{APPLY}(T, \hat{A}, e, \theta),$$

where for an effect specification $e : [q^{++}] \wedge q^= \wedge Q^- \rightsquigarrow \hat{A}'$, we have

$$\text{APPLY}(T, \hat{A}, e, \theta) = \bigcup_{\sigma \in \text{ANS}(q^{++}\theta, \emptyset, \hat{A}) \bowtie \text{ANS}((q^= \wedge Q^-)\theta, \emptyset, \hat{A})} \hat{A}'\theta\sigma \;\; \cup \text{EQ}(T, \hat{A}).$$

Instead, we define $\text{DO}_{\text{NORM}}()$ as

$$\text{DO}_{\text{NORM}}(T, \hat{A}, \hat{\gamma}\theta) = \bigcup_{e \text{ in } \hat{\gamma}} \text{APPLY}_{\text{NORM}}(T, \hat{A}, e, \theta),$$

where, for an effect specification $e : [q^{++}] \wedge q^= \wedge Q^- \rightsquigarrow \hat{A}'$, we have

$$\text{APPLY}_{\text{NORM}}(T, \hat{A}, e, \theta) = \bigcup_{\sigma \in \text{ANS}(q^{++}\theta, \emptyset, \hat{A}^{\not\models Q}) \bowtie \text{ANS}((q^= \wedge Q^-)\theta, \emptyset, \hat{A})} \hat{A}'\theta\sigma \quad \cup \text{EQ}(T, \hat{A}).$$

Notice that the only difference between $\text{DO}()$ and $\text{DO}_{\text{NORM}}()$ is that in the latter we use only $\hat{A}^{\not\models Q}$ instead of $\hat{A}$ to compute the answers to the CQs $q^{++}\theta$.

The following lemma shows that the applications of $\text{DO}()$ and of $\text{DO}_{\text{NORM}}()$ give rise to logically equivalent ABoxes.

**Lemma 10** $\text{DO}(T, \hat{A}, \hat{\gamma}\theta) \equiv_{T, \text{ALPH}(\mathcal{K})} \text{DO}_{\text{NORM}}(T, \hat{A}, \hat{\gamma}\theta)$.

*Proof.* In order to prove the claim, it is enough to show that for each concept/role assertion $\alpha_2 \in \text{DO}_{\text{NORM}}(T, \hat{A}, \hat{\gamma}\theta)$ whose concept/role name belongs to $\text{ALPH}(\mathcal{K})$, we have that $(T, \text{DO}(T, \hat{A}, \hat{\gamma}\theta)) \models \alpha_2$, and for each concept/role assertion $\alpha_1 \in \text{DO}(T, \hat{A}, \hat{\gamma}\theta)$ whose concept/role name belongs to $\text{ALPH}(\mathcal{K})$, we have that $(T, \text{DO}_{\text{NORM}}(T, \hat{A}, \hat{\gamma}\theta)) \models \alpha_1$. We actually prove a slightly stronger result:

*(1)* For each ABox assertion $\alpha_2 \in \text{APPLY}_{\text{NORM}}(T, \hat{A}, e, \theta)$, we have that $(T, \text{APPLY}(T, \hat{A}, e, \theta)) \models \alpha_2$.

*(2)* For each ABox assertion $\alpha_1 \in \text{APPLY}(T, \hat{A}, e, \theta)$, we have that $(T, \text{APPLY}_{\text{NORM}}(T, \hat{A}, e, \theta)) \models \alpha_1$.

For *(1)*, by monotonicity of $q^{++}$ and the fact that $\hat{A}^{\not\models Q} \subseteq \hat{A}$, we have that

$$\bigcup_{\sigma \in (\text{ANS}(q^{++}\theta, \emptyset, \hat{A}^{\not\models Q}) \bowtie \text{ANS}((q^= \wedge Q^-)\theta, \emptyset, \hat{A}))} \hat{A}'\theta\sigma \quad \text{is contained in} \quad \bigcup_{\sigma \in (\text{ANS}(q^{++}\theta, \emptyset, \hat{A}) \bowtie \text{ANS}((q^= \wedge Q^-)\theta, \emptyset, \hat{A}))} \hat{A}'\theta\sigma,$$

hence the claim follows.

For *(2)*, consider an ABox assertion $\alpha \in \text{APPLY}(T, \hat{A}, e, \theta)$. By definition of $\text{APPLY}()$, we know that there exists an effect $e : [q^{++}] \wedge q^= \wedge Q^- \rightsquigarrow \hat{A}'$ and an assignment $\sigma$ to the free variables of $q^{++}$ (which include also the free variables of $q^= \wedge Q^-$) such that $\sigma \in (\text{ANS}(q^{++}\theta, \emptyset, \hat{A}) \bowtie \text{ANS}((q^= \wedge Q^-)\theta, \emptyset, \hat{A}))$ and $\alpha \in \hat{A}'\theta\sigma$. Let $\{x_1, \ldots, x_n\}$ be all free variables in $q^{++}\theta$, and $\sigma = \{x_1 \to t_1, \ldots, x_n \to t_n,\}$. For each variable $x_i$, let $N(x_i)$ be the (unique) concept atom in $q^{++}\theta$ in which $x_i$ occurs (similar considerations hold when $x_i$ occurs in a role atom). Then, either $N(t_i) \in \hat{A}^{\not\models Q}$, or for some $t'_i$, $N(t'_i) \in \hat{A}^{\not\models Q}$ and $(t_i = t'_i) \in \text{EQ}(T, \hat{A})$. In the former case, let $t''_i$ denote $t_i$, while in the latter case let $t''_i$ denote $t'_i$. Then, consider the substitution $\sigma' = \{x_1 \to t''_1, \ldots, x_n \to t''_n,\}$. By construction, we have that $\sigma' \in \text{ANS}(q^{++}\theta, \emptyset, \hat{A}^{\not\models Q})$, and since $\sigma \in \text{ANS}((q^= \wedge Q^-)\theta, \emptyset, \hat{A})$, and $(t''_i = t_i) \in \text{EQ}(T, \hat{A})$ for each $i \in \{1, \ldots, n\}$, we also have that $\sigma' \in \text{ANS}(q^{++}\theta, \emptyset, \hat{A}^{\not\models Q}) \bowtie \text{ANS}((q^= \wedge Q^-)\theta, \emptyset, \hat{A})$. Since

- $\alpha \in \hat{A}'\theta\sigma$,

- $\sigma$ and $\sigma'$ are identical modulo $\text{EQ}(T, \hat{A})$ and
- $\text{EQ}(T, \hat{A}) \subseteq \text{APPLY}_{\text{NORM}}(T, \hat{A}, e, \theta)$,

we can infer that $(T, \text{APPLY}_{\text{NORM}}(T, \hat{A}, e, \theta)) \models \alpha$. Hence the claim holds. □

By combining Lemma 8 and Lemma 10, we get that $\text{DO}()$ on $\mathcal{K}$ and $\text{DO}_{\text{NORM}}()$ on $\hat{\mathcal{K}}$ behave equivalently, when starting from equivalent ABoxes.

**Lemma 11** *If $A_1 \equiv_{T,\text{ALPH}(\mathcal{K})} A_2$ then $\text{DO}(T, A_1, \gamma\theta) \equiv_{T,\text{ALPH}(\mathcal{K})} \text{DO}_{\text{NORM}}(T, A_2, \hat{\gamma}\theta)$.*

*Proof.* The claim is a direct consequence of Lemma 8, Lemma 10, the equivalence between $A_1$ and $A_2$, and the observation that logical equivalence is transitive. □

Given a KAB $\mathcal{K}$ and its normalized version $\hat{\mathcal{K}}$, we call the transition system generated in the same way as $\Upsilon_\mathcal{K}$, but using $\text{DO}_{\text{NORM}}()$ on $\hat{\mathcal{K}}$ instead of $\text{DO}()$ on $\mathcal{K}$, the *normalized transition system* generated by $\hat{\mathcal{K}}$, and denote it with $\Upsilon_{\hat{\mathcal{K}},\text{NORM}}$.

**Lemma 12** *Given a KAB $\mathcal{K}$, the transition systems $\Upsilon_\mathcal{K}$ and $\Upsilon_{\hat{\mathcal{K}},\text{NORM}}$ are bisimilar.*

*Proof.* Let $\Upsilon_\mathcal{K} = (\mathbb{U}, T, \Sigma, s_0, abox, \Rightarrow)$ and $\Upsilon_{\hat{\mathcal{K}},\text{NORM}} = (\mathbb{U}, T, \Sigma_{\text{NORM}}, s_0, abox_{\text{NORM}}, \Rightarrow_{\text{NORM}})$. We define the relation $\mathcal{B} \subseteq \Sigma \times \Sigma_{\text{NORM}}$ as follows: $(s_1, s_2) \in \mathcal{B}$ iff $abox(s_1) \equiv_{T,\text{ALPH}(\mathcal{K})} abox_{\text{NORM}}(s_2)$ and show that $\mathcal{B}$ is a bisimulation. To do so, we prove that $\mathcal{B}$ is closed under the definition of bisimulation itself. Indeed, if $(s_1, s_2) \in \mathcal{B}$, then:

- $abox(s_1) \equiv_{T,\text{ALPH}(\mathcal{K})} abox(s_2)$ by definition.
- If $s_1 \Rightarrow s'_1$ then there exists an action $\gamma$ and a substitution $\theta$ such that $s'_1 = \text{DO}(T, abox(s_1), \gamma\theta)$ (notice that $abox(s_1) = s_1$) and $s'_1$ is consistent w.r.t. $T$. Now let us consider $s'_2 = \text{DO}_{\text{NORM}}(T, abox(s_2), \hat{\gamma}\theta)$. Since $abox(s_1) \equiv_{T,\text{ALPH}(\mathcal{K})} abox(s_2)$, then by Lemma 11, we have $s'_1 \equiv_{T,\text{ALPH}(\mathcal{K})} s'_2$. Therefore, $s'_2$ is consistent w.r.t. $T$, and hence $s_2 \Rightarrow_{\text{NORM}} s'_2$, and $(s'_1, s'_2) \in \mathcal{B}$.
- Similarly, if $s_2 \Rightarrow_{\text{NORM}} s'_2$ then there exists an action $\hat{\gamma}$ and a substitution $\theta$ such that $s'_2 = \text{DO}_{\text{NORM}}(T, abox(s_2), \hat{\gamma}\theta)$ and $s'_2$ is consistent w.r.t. $T$. Now let us consider $s'_1 = \text{DO}(T, abox(s_1), \gamma\theta)$. Since $s_2 \equiv_{T,\text{ALPH}(\mathcal{K})} s_1$, then by by Lemma 11, we have $s'_2 \equiv_{T,\text{ALPH}(\mathcal{K})} s'_1$ Therefore, $s'_1$ is consistent w.r.t. $T$, and hence $s_1 \Rightarrow s'_1$, and, considering that equivalence enjoys symmetry, we have $(s'_1, s'_2) \in \mathcal{B}$.

This proves the claim. □

The direct consequence of the above lemma is that, by considering the Bisimulation Invariance Theorem 4, we can faithfully check $\mu\mathcal{L}_A$ formulas over $\Upsilon_{\hat{\mathcal{K}},\text{NORM}}$ instead of $\Upsilon_\mathcal{K}$.

### 7.3 Positive Dominant

Our next step is to show that for a weakly acyclic KAB $\mathcal{K}$, the normalized transition system $\Upsilon_{\hat{\mathcal{K}},\text{NORM}}$ is finite. We do so by considering another transition system, which is behaviorally unrelated to $\Upsilon_{\hat{\mathcal{K}},\text{NORM}}$, and hence to $\Upsilon_\mathcal{K}$, but whose active domain bounds the active domain of $\Upsilon_{\hat{\mathcal{K}},\text{NORM}}$. We obtain such a transition system essentially by ignoring all negative information and equalities. This allows us to refer back to the literature on data exchange to show boundedness. We call such a transition system *positive dominant*.

Given a normalized KAB $\hat{\mathcal{K}} = (T, \hat{A}_0, \hat{\Gamma}, \Pi)$, we define the *positive dominant* of $\mathcal{K}$ as the KAB

$$\mathcal{K}^+ = (\emptyset, \hat{A}_0^{\neq Q}, \{\gamma^+\}, \{\text{true} \mapsto \gamma^+\}).$$

The only action $\gamma^+$ is without parameters and its effect specification is constituted by *CopyAll* and by one effect of the form

$$[q_i^{++}] \rightsquigarrow A_i'^{\not\models Q}$$

for each effect $[q_i^{++}] \wedge q_i^= \wedge Q_i^- \rightsquigarrow A_i'$ in every action of $\hat{\Gamma}$. Observe that the parameters of the actions in $\hat{\Gamma}$ become simply free variables in $\gamma^+$.

Notice that $\gamma^+$ is applicable at every step because the process trivially always allows it. The resulting state is always consistent, since $\mathcal{K}^+$ has an empty TBox. Moreover, no equality assertion is ever generated. The transition system $\Upsilon_{\mathcal{K}^+}$ is constituted by a single run, which incrementally accumulates all the facts that can be derived by the iterated application of $\gamma^+$ over such increasing ABox. This behavior closely resembles the chase of tuple-generating dependencies (TGDs) in data exchange, where an application of $\gamma^+$ corresponds to a "parallel" chase step (Deutsch, Nash, & Remmel, 2008).

From a technical point of view, notice that $\mathcal{K}^+$ is already in normalized form (i.e., $\mathcal{K}^+ = \hat{\mathcal{K}^+}$), and that DO() and DO$_{\text{NORM}}$() are identical since neither equality nor negation are considered. Hence $\Upsilon_{\mathcal{K}^+} = \Upsilon_{\hat{\mathcal{K}^+},\text{NORM}}$.

The next lemma shows that $\mathcal{K}^+$ preserves weak acyclicity of $\hat{\mathcal{K}}$.

**Lemma 13** *If $\hat{\mathcal{K}}$ is weakly acyclic then also its positive dominant $\mathcal{K}^+$ is weakly acyclic.*

*Proof.* The claim follows from the fact that, by construction, the dependency graph $\mathcal{G}^+$ of $\mathcal{K}^+$ is equal to $\hat{\mathcal{G}}$. Indeed, both $q_i^{++}$ and its connection with $\hat{A}_i$ are preserved by $\mathcal{K}^+$. Hence, we get the claim. □

Next we show that if $\mathcal{K}^+$ is weakly acyclic the active domain of the ABoxes in its transition system $\Upsilon_{\mathcal{K}^+}$ are polynomially bounded by the active domain of the initial ABox.

**Lemma 14** *If $\mathcal{K}^+$ is weakly acyclic, then there exists a polynomial $\mathcal{P}(\cdot)$ such that*

$$|\text{ADOM}(\Upsilon_{\mathcal{K}^+})| < \mathcal{P}(|\text{ADOM}(\hat{A}_0^{\not\models Q})|).$$

*Proof.* We observe that there exists a strict connection between the execution of $\mathcal{K}^+$ and the chase of a set of TGDs in data exchange. Therefore, the proof closely resembles the one by Fagin et al. (2005, Thm. 3.9), where it is shown that for weakly acyclic TGDs, every chase sequence is bounded.

Let $\Upsilon_{\mathcal{K}^+} = (\mathbb{U}, \emptyset, \Sigma, A_0^{\not\models Q}, abox, \Rightarrow)$, let $\mathcal{G}^+ = (V, E)$ be the dependency graph of $\mathcal{K}^+$, and let $n = |\text{ADOM}(A_0^{\not\models Q})|$. For every node $p \in V$, we consider an *incoming path* to be any (finite or infinite) path ending in $p$. Let $rank(p)$ be the maximum number of special edges on any such incoming path. Since $\mathcal{K}^+$ is weakly acyclic by hypothesis, $\mathcal{G}^+$ does not contain cycles going through special edges, and therefore $rank(p)$ is finite. Let $r$ be the maximum among $rank(p_i)$ over all nodes. We observe that $r \leq |V|$; indeed no path can lead to the same node twice using special edges, otherwise $\mathcal{G}^+$ would contain a cycle going through special edges, thus breaking the weak acyclicity hypothesis. Next we observe that we can partition the nodes in $V$ according to their rank, obtaining a set of sets $\{V_0, V_1, \ldots, V_r\}$, where $V_i$ is the set of all nodes with rank $i$.

Let us now consider a state $A$ obtained from $A_0^{\not\models Q}$ by applying the only action $\gamma^+$ contained in $\mathcal{K}^+$ an arbitrary number of times. We now prove, by induction on $i$, the following claim: for every $i$ there exists a polynomial $\mathcal{P}_i$ such that the total number of distinct values $c$ that occur in $A$ at positions in $V_i$ is at most $\mathcal{P}_i(n)$.

**(Base case)** Consider $p \in V_0$. By definition, $p$ has no incoming path containing special edges. Therefore, no new values are stored in $p$ along the run $A_0^{\not\exists Q} \Rightarrow \cdots \Rightarrow A$. Indeed $p$ can just store values that are part of the initial ABox $A_0^{\not\exists Q}$. This holds for all nodes in $V_0$ and hence we can fix $\mathcal{P}_0(n) = n$.

**(Inductive step)** Consider $p \in V_i$, with $i \in \{1, \ldots, r\}$. The first kind of values that may be stored inside $p$ are those values that were stored inside $p$ itself in $A_0^{\not\exists Q}$. The number of such values is at most $n$. In addition, a value may be stored in $p$ for two reasons: either it is copied from some other position $p' \in V_j$ with $i \neq j$, or it is generated as a possibly new function term, built when applying effects that contain a function in their head.

We first determine the number of fresh individuals that can be generated from function terms. The possibility of generating and storing a new value in $p$ as a result of an action is reflected by the presence of special edges. By definition, any special edge entering $p$ must start from a node $p' \in V_0 \cup \cdots \cup V_{i-1}$. By induction hypothesis, the number of distinct values that can exist in $p'$ is bounded by $\mathcal{H}(n) = \sum_{j \in \{0, \ldots, i-1\}} \mathcal{P}_j(n)$. Let $b_a$ be the maximum number of special edges that enter a position, over all positions in the TBox; $b_a$ bounds the arity taken by each function term contained in $\gamma$. Then for every choice of $b_a$ values in $V_0 \cup \cdots \cup V_{i-1}$ (one for each special edge that can enter a position), the number of new values generated at position $p$ is bounded by $t_f \cdot \mathcal{H}(n)^{b_a}$, where $t_f$ is the total number of facts contained in all effects of $\gamma^+$. Note that this number does not depend on the data in $A_0^{\not\exists Q}$. By considering all positions in $V_i$, the total number of values that can be generated is then bounded by $\mathcal{F}(n) = |V_i| \cdot t_f \cdot \mathcal{H}(n)^{b_a}$. Clearly, $\mathcal{F}(\cdot)$ is a polynomial, because $t_f$ and $b_a$ are determined by $\gamma^+$.

We count next the number of distinct values that can be copied to positions of $V_i$ from positions of $V_j$, with $j \neq i$. A copy is represented in the graph as a normal edge going from a node in $V_j$ to a node in $V_i$, with $j \neq i$. We observe first that such normal edges can start only from nodes in $V_0 \cup \cdots \cup V_{i-1}$, that is, they cannot start from nodes in $V_j$ with $j > i$. We prove this by contradiction. Assume that there exists $p' \to p \in E$, such that $p \in V_i$ and $p' \in V_j$ with $j > i$. In this case, the rank of $p$ would be $j > i$, which contradicts the fact that $p \in V_i$. As a consequence, the number of distinct values that can be copied to positions in $V_i$ is bounded by the total number of values in $V_0 \cup \cdots \cup V_{i-1}$, which corresponds to $\mathcal{H}(n)$ from our previous consideration.

Putting it all together, we define $\mathcal{P}_i(n) = n + \mathcal{F}(n) + \mathcal{H}(n)$. Since $\mathcal{P}_i(\cdot)$ is a polynomial, the claim is proven.

Notice that, in the above claim, $i$ is bounded by $r$, which is a constant. Hence, there exists a fixed polynomial $\mathcal{P}(\cdot)$ such that the number of distinct values that can exist in every state $s \in \Sigma$ is bounded by $\mathcal{P}(n)$. $\mathcal{K}^+$ is inflationary, because when $\gamma^+$ is applied it copies all concept and role assertions from the current to the next state. Since $\Upsilon_{\mathcal{K}^+}$ contains only a single run, $\mathcal{P}(n)$ is a bound for $\text{ADOM}(\Upsilon_{\mathcal{K}^+})$ as well. $\square$

The following lemma shows the key feature of the positive dominant.

**Lemma 15** $\text{ADOM}(\Upsilon_{\hat{\mathcal{K}}}) \subseteq \text{ADOM}(\Upsilon_{\mathcal{K}^+})$.

*Proof.* Let $\hat{\mathcal{K}} = (T, \hat{A}_0, \hat{\Gamma}, \Pi)$ and $\mathcal{K}^+ = (\emptyset, \hat{A}_0^{\not\exists Q}, \{\gamma^+\}, \{\text{true} \mapsto \gamma^+\})$.

We first observe that, for every ABox $A$ in $\Upsilon_{\hat{\mathcal{K}}}$, $\text{ADOM}(A) = \text{ADOM}(A^{\not\exists Q})$ by definition of $\hat{\mathcal{K}}$ (this is the role of the special concept $Dummy$).

We show by induction on the construction of $\Upsilon_{\hat{\mathcal{K}}}(\mathbb{U}, T, \Sigma_1, \hat{A}_0, abox, \Rightarrow_1)$ and $\Upsilon_{\mathcal{K}^+} = (\mathbb{U}, \emptyset, \Sigma_2, \hat{A}_0^{\not\models Q}, abox, \Rightarrow_2)$, that for each state $A_1 \in \Sigma_1$ we have that there exists a state $A_2 \in \Sigma_2$ such that $A_1^{\not\models Q} \subseteq A_2$.

The base case holds for the initial states $\hat{A}_0$ and $\hat{A}_0^{\not\models Q}$ of the two transition systems by definition. For the inductive case, we have to show that, given $A_1 \in \Sigma_1$ and $A_2 \in \Sigma_2$ with $A_1^{\not\models Q} \subseteq A_2$, for each $A_1' \in \Sigma_1$ with $A_1 \Rightarrow_1 A_1'$, the unique state $A_2' \in \Sigma_2$ with $A_2 \Rightarrow_2 A_2'$ is such that $A_1' \subseteq A_2'$. To show this, note that $A_1 \Rightarrow_1 A_1'$ if there exists an action $\gamma$ of $\hat{\mathcal{K}}$ and a substitution $\theta$ for the parameters of $\gamma$ such that $A_1' = \text{DO}_{\text{NORM}}(T, A_1, \gamma\theta)$. Similarly, taking into account that $\gamma^+$ has no parameters and is always executable in $\Upsilon_{\mathcal{K}^+}$, we have that $A_2' = \text{DO}(T, A_2, \gamma^+) = \text{DO}_{\text{NORM}}(T, A_2, \gamma^+)$. By construction of $\mathcal{K}^+$, for each effect $e_1 \in \gamma$ of the form

$$e_1 : [q^{++}] \wedge q^= \wedge Q^- \rightsquigarrow A'_{e_1},$$

there is an effect $e_2 \in \gamma^+$ of the form

$$e_2 : [q^{++}] \rightsquigarrow A'_{e_1}{}^{\not\models Q},$$

where $A'_{e_1}{}^{\not\models Q}$ is obtained from $A'_{e_1}$ by removing all equality assertions. By induction hypothesis, we have that $A_1^{\not\models Q} \subseteq A_2$. By observing that $\text{ANS}([q^{++}]\theta, \emptyset, A_1^{\not\models Q}) \bowtie \text{ANS}((q^= \wedge Q^-)\theta, \emptyset, A_1) \subseteq \text{ANS}([q^{++}], \emptyset, A_2)$, we then obtain that $A'_{e_1}{}^{\not\models Q} \subseteq A'_{e_2}$, where $A'_{e_1} = \text{APPLY}_{\text{NORM}}(T, A_1, e_1, \theta)$ and $A'_{e_2} = \text{APPLY}(\emptyset, A_2, e_2, \emptyset)$. Hence, we get the claim that $A_1'^{\not\models Q} \subseteq A_2'$.

Now since for an ABox $A$ of $\Upsilon_{\hat{\mathcal{K}}}$ the active domain $\text{ADOM}(A)$ of $A$ and $\text{ADOM}(A^{\not\models Q})$ are identical by construction, and since $\text{ADOM}(\Upsilon_{\hat{\mathcal{K}}})$ and $\text{ADOM}(\Upsilon_{\mathcal{K}^+})$ are simply the union of the active domains of all generated ABoxes, we get the claim. □

### 7.4 Putting it All Together

If a KAB $\mathcal{K}$ is weakly acyclic, then, by Lemma 9, its normalized form $\hat{\mathcal{K}}$ is weakly acyclic as well and, by Lemma 13, so is its positive dominant $\mathcal{K}^+$. Hence, by Lemma 14, the size of the active domain $\text{ADOM}(\Upsilon_{\mathcal{K}^+})$ of the transition system $\Upsilon_{\mathcal{K}^+}$ of $\mathcal{K}^+$ is polynomially related to the size of its initial ABox.

Now, by Lemma 15, this implies that also the size of the active domain $\text{ADOM}(\Upsilon_{\hat{\mathcal{K}},\text{NORM}})$ of the transition system $\Upsilon_{\hat{\mathcal{K}}}$ of $\hat{\mathcal{K}}$ is polynomially related to the size of its initial ABox. Hence, the number of possible states of $\Upsilon_{\hat{\mathcal{K}}}$ is finite, and in fact at most exponential in the size of the initial ABox. It follows that checking $\mu\mathcal{L}_A$ formulae over $\Upsilon_{\hat{\mathcal{K}}}$ can be done in EXPTIME w.r.t. the size of $\mathcal{K}$.

Finally, by Lemma 12, $\Upsilon_{\hat{\mathcal{K}}}$ and $\Upsilon_{\mathcal{K}}$ are bisimilar, and by the Bisimulation Invariance Theorem 4, $\Upsilon_{\hat{\mathcal{K}}}$ and $\Upsilon_{\mathcal{K}}$ satisfy exactly the same $\mu\mathcal{L}_A$ formulae. Hence, to check a $\mu\mathcal{L}_A$ formula on $\Upsilon_{\mathcal{K}}$ it is sufficient to check it over $\Upsilon_{\hat{\mathcal{K}}}$, which can be done in EXPTIME. This concludes the proof of Theorem 7. □

## 8. Related Work

We provide now a detailed review of work that is related to the framework and the results presented in the previous sections.

### 8.1 Combining Description Logics and Temporal Logics

Our work is deeply related to the research that studies combinations of description logics and temporal logics. Indeed, actions progress knowledge over time and, although temporal logics do not mention actions, we can easily used them for describing progression mechanisms, including transition systems (see, e.g., Clarke et al., 1999; Calvanese, De Giacomo, & Vardi, 2002).

Such research has mostly explored the combination of standard description logics with standard temporal logics at the level of models, which is certainly the most natural form of combination from a logical point of view. Technically, this form of combination gives rise to a combined logic with a two-dimensional semantics, where one dimension is for time and the other for the DL domain (Schild, 1993; Wolter & Zakharyaschev, 1999b, 1999a; Gabbay et al., 2003). Unfortunately, from a computational point of view, this form of combination suffers from a key undecidability result, which makes it too fragile for many practical purposes: the possibility of specifying that roles preserve their extension over time (the so called *rigid roles*) causes undecidability[6]. Referring to the domain of interest in Example 1, this would result, for example, in the undecidability of theories that specify that each instance of Character livesIn the same City forever. Moreover, this undecidability result already holds for concept satisfiability w.r.t. a fixed TBox (i.e., where the same TBox axioms must hold at all time points), without ABoxes, and with only a single rigid role (Wolter & Zakharyaschev, 1999b, 1999a; Gabbay et al., 2003). That is, it holds for a reasoning service that is much simpler than conjunctive query answering (Calvanese, De Giacomo, & Lenzerini, 2008), even with a fixed TBox and no data (no ABox assertions, hence no individual terms) and for one of the simplest kinds of temporal formulae, namely "forever something is true" (safety) (Clarke et al., 1999).

Decidability can be regained by: *(i)* dropping TBoxes altogether, but the decision problem is still hard for non-elementary time (Gabbay et al., 2003); *(ii)* allowing temporal operators only on concepts (Schild, 1993; Artale & Franconi, 1998, 2005; Gutiérrez-Basulto, Jung, & Lutz, 2012; Jamroga, 2012), and in this case the complexity depends crucially on the description logic; *(iii)* allowing temporal operators only on TBox and ABox assertions (Lutz, Wolter, & Zakharyaschev, 2008; Baader et al., 2012). In fact cases *(ii)* and *(iii)* can be mixed (Baader & Laux, 1995; Wolter & Zakharyaschev, 1998).

Allowing for temporal operators over assertions only (case *(iii)* above), is tightly related to the functional approach adopted in this paper: the fact that we admit temporal operators only in front of assertions allows us to consider temporal models whose time points are actually sets of models of description logic assertions. Hence it keeps the temporal component distinct from the description logic one, exactly as we do here. In particular, the results by Baader et al. (2012) can be directly compared with ours. Apart from the obvious differences in the formalism used, one key point to get decidability there is that the individual terms mentioned in the ABox assertions are fixed a priori. It is possible that, by adapting the techniques presented here, those results could be extended to allow functions for denoting terms, hence allowing for adding fresh individual terms during the temporal evolution.

---

6. To lose decidability, it suffices to be able to specify/verify the persistence of binary predicates/roles, which allows one to build an infinite grid and hence to encode any Turing-machine computation (Robinson, 1971; van Emde Boas, 1997).

## 8.2 Combining Description Logics and Actions

Somehow hampered by the undecidability results mentioned at the beginning of the section, also combinations of description logics and action theories have been studied in the years. In particular, Liu, Lutz, Milicic, and Wolter (2006b, 2006a) study combinations of description logics and action theories at the level of models, but only w.r.t. the two classical problems in reasoning about actions, namely *projection* and *executability*. Both of these problems require to explicitly give a sequence of actions and then check a property of the resulting final state (projection), or check the executability of the sequence of actions, each of which comes with a certain precondition (Reiter, 2001). More sophisticated temporal properties (in particular, "forever something is true" mentioned above) would lead to undecidability. By the way, notice that such undecidability result also deeply questions from the computational point of view the possibility of adding (sound and complete) automated reasoning capabilities to proposals such as OWL-S (Semantic Markup for Web Services) (Martin, Paolucci, McIlraith, Burstein, McDermott, McGuinness, Parsia, Payne, Sabou, Solanki, Srinivasan, & Sycara, 2004).

Possibly the first proposal based implicitly on the functional view of the KB was the pioneering work by De Giacomo, Iocchi, Nardi, and Rosati (1999), which adopts an epistemic description logic (based on certain answers) combined with an action formalism to describe routines of a mobile robot. Again, one important point there is that individual terms are bounded and fixed a priori. The functional view approach was first spelled out by Calvanese, De Giacomo, Lenzerini, and Rosati (2007), and by Calvanese et al. (2011). In that work, only projection and executability are studied, however there is a distinction between the KB in the states and the actions (there specified as updates), so that the framework gives rise to a single transition system whose states are labeled by KBs (in fact the TBox is fixed while the ABox changes from state to state). However, again, the individual terms considered are fixed a priori and hence the resulting transition system is finite. So, although not studied in that work, sophisticated forms of temporal properties as those proposed here are readily verifiable in that setting. Interestingly, apart from the KBs and action, in that work also Golog-like programs are considered. These are programs whose atomic actions are defined by the action formalism, and are combined using (usual and less usual) programming constructs, such as sequence, while-loop, if-then-else, and nondeterministic pick of a value (Levesque, Reiter, Lesperance, Lin, & Scherl, 1997; De Giacomo, Lespérance, & Levesque, 2000). An important characteristic of these programs is that they have a finite number of control states (notice that the memory storage of these programs is kept in the action theory, or the KB in our case). Although out of the scope of this paper, this finiteness allows for easily extending our results to such program as well.

An interesting alternative way to combine description logics and reasoning about actions is the one reported by Gu and Soutchanski (2010). There, a description logics KB[7] is used as a special FOL theory describing the initial situation in a situation calculus basic action theory (Reiter, 2001). Notice that as a result, TBox assertions do not act as state constraints (Lin & Reiter, 1994), which would lead to undecidability as discussed above (Wolter & Zakharyaschev, 1999b, 1999a; Gabbay et al., 2003), in fact they essentially do not persist in any way through actions.

---

7. They actually mainly focus on concepts only but in a description logic that includes the universal role, which allows one to express TBox assertions as concepts (Baader et al., 2003).

### 8.3 Description Logics Update

Observe that effects of an action in our setting can be seen as a basic form of update of the previous state (Katsuno & Mendelzon, 1991). Although our mechanism sidesteps the semantic and computational difficulties of description logic KB update (Liu et al., 2006b; De Giacomo, Lenzerini, Poggi, & Rosati, 2009; Calvanese, Kharlamov, Nutt, & Zheleznyakov, 2010; Lenzerini & Savo, 2012) by simply rejecting the execution of actions that would lead to an inconsistent state. Adopting proper forms of update in our setting is an interesting issue for future research.

### 8.4 Artifacts and Data-Aware Processes

Our work is also closely related to research in verification of artifact-centric business processes (Nigam & Caswell, 2003; Bhattacharya et al., 2007). Artifact-centric approaches model business processes by giving equal importance to the control-flow perspective and the data of interest. An artifact is typically represented as a tuple of a schema, which models the artifact type, together with a set of actions/services that specify how the information maintained in the artifact can be manipulated over time. Each action is usually represented in terms of pre- and post-conditions that are respectively used to determine when the action is eligible for execution, and to relate the current artifact state with the successor state obtained after the action execution. Pre- and post-conditions are modeled as first-order formulae, and post-conditions employ existentially quantified variables to account for external inputs from the environment. Differently from KABs, most of the approaches targeting artifact-centric processes assume complete information about data, using a relational database to maintain the artifacts' information. As in this paper, the aim of such works is to verify whether a relational artifact-centric process meets some temporal/dynamic property, formalized using first-order variants of branching or linear temporal logics.

In the work by Deutsch et al. (2009), the infinite domain of the artifact's database is equipped with a dense linear order, which can be mentioned in pre-conditions, post-conditions, and properties. Runs can receive unbounded external input from an infinite domain. Decidability of verification is achieved by avoiding branching time properties, and by restricting the formulae used to specify pre-, post-conditions and properties. In particular, the approach refers to read-only and read-write database relations differently, querying the latter only by checking whether they contain a given tuple of constants. The authors show that this restriction is tight, and that integrity constraints cannot be added to the framework, since even a single functional dependency leads to undecidability of verification. Damaggio et al. (2011) extend this approach by disallowing read-write relations, but this allows the extension of the decidability result to integrity constraints expressed as embedded dependencies with terminating chase, and to any decidable arithmetic. This is a major difference with our approach, where all concepts of the KAB are considered as read-write relations, and can be arbitrarily queried to determine the progression of the system. Differently from these works, Belardinelli et al. (2011) consider a first-order variant of CTL with no quantification across states as verification formalism. The framework supports the incorporation of new values from the external environment as parameters of the actions; the corresponding execution semantics considers all the possible actual values, thus leading to an infinite-state transition systems. As for decidability of verification, the authors show that, under the assumption that each state of the system (constituted by the union of artifacts' relational instances) has a bounded active domain, it is possible to construct a faithful abstract transition system which, differently from the original one, has a finite number of states. Belardinelli, Lomuscio, and Patrizi (2012) improve the results by Belar-dinelli et al. (2011) by introducing a semantic property of "uniformity" which, roughly speaking, says that the transition system representing the execution of the process under study is not able to distinguish among states that have the same constants and the same patterns of data. Under the assumptions of uniformity and state boundedness, decidability of verification is achieved for a richer logic, namely CTL with quantification across states, interpreted under the active domain semantics. The notion of state boundedness has also been adopted by the independently developed framework of Bagheri Hariri, Calvanese, De Giacomo, Deutsch, and Montali (2012, 2013), where first-order variants of $\mu$-calculus, similar to the one considered here, are considered. There, beside differences in the way data and external information are modeled, sufficient syntactic conditions that guarantee state boundedness are proposed. All these works are developed within the relational database setting, and do not extend trivially to systems where actions change DL knowledge bases.

The connection between data-/artifact-centric business processes and data exchange that we exploit in this paper was first established by Cangialosi et al. (2010), and by De Giacomo, De Masellis, and Rosati (2012). There the transition relation itself is described in terms of TGDs, which map the current state, represented as a relational database instance, to the next one. Null values are used to model the incorporation of new, unknown data into the system. The process evolution is essentially a form of chase. Under suitable weak acyclicity conditions this chase terminates, guaranteeing, in turn, that the system is finite-state. Decidability is then shown for a first-order $\mu$-calculus without first-order quantification across states. This approach was extended by Bagheri Hariri et al. (2011), where TGDs were replaced by actions and a rule-based process that follow the same structure of the KAB action component. In this revised framework, values imported from the external environment are represented by uninterpreted function terms, which play the same role as nulls in the work by Cangialosi et al. (2010), and by De Giacomo et al. (2012). Since Bagheri Hariri et al. (2011), Cangialosi et al. (2010), and De Giacomo et al. (2012) all rely on a purely relational setting, this choice leads to an ad-hoc interpretation of equality, where each null value/function term is considered only equal to itself. Differently from these works, here we allow for sophisticated schema constraints, i.e., the TBox itself, and provide at the same time a more fine-grained treatment of equality, where individuals can be inferred to be equal due to the application of such schema constraints and/or the execution of some action. This treatment of equality differentiates this work also from the one of Bagheri Hariri, Calvanese, De Giacomo, and De Masellis (2011), which introduces a preliminary version of the framework here presented, where UNA is assumed and equality is not considered. More specifically, Bagheri Hariri et al. (2011) propose *semantic artifacts* as a means to represent artifacts and corresponding processes at a higher level of abstraction than relational artifacts, representing the artifact data with a semantically rich knowledge base operating with incomplete information. KABs constitute a more general framework, which can be seamlessly customized to account for semantic artifacts. A major difference with the work by Bagheri Hariri et al. (2011) is also constituted by the verification formalism. In particular, both works focus on a form of $\mu$-calculus where ECQs are used to query the states of the system, but Bagheri Hariri et al. (2011) do not support quantification across states, as done here.

Calvanese et al. (2012) investigate a framework for data-centric processes that mixes the approach proposed by Bagheri Hariri et al. (2013) for relational artifacts with the notion of knowledge bases as used here. In particular, semantically-governed data-aware processes are introduced as a mechanism to model a dynamic system working over a relational database, providing at the same time a conceptual representation of the manipulated data in terms of a *DL-Lite* knowledge base. By relying on ontology-based data access (Calvanese et al., 2009), declarative mappings are used to connect the knowledge base with the underlying relational database. Differently from KABs, the system evolves at the relational layer, and the knowledge base is used to understand and ultimately govern such an execution at a higher level of abstraction.

We observe that the results presented here fully subsume those by Bagheri Hariri et al. (2011), where the underlying description logic is OWL 2 QL. On the one hand, if we remove the possibility of asserting functionality of roles in the knowledge component, and of equating individuals as a result of an action in the action component, we precisely obtain the setting presented by Bagheri Hariri et al. (2011). On the other hand, for both frameworks the established complexity upper bounds are the same.

## 9. Conclusions

In this paper we have studied verification of knowledge and action bases, which are dynamic systems constituted by a knowledge base, expressed in description logics, and by an action specification that changes the knowledge base over time. We have obtained an interesting decidability result by relying on the notion of weak acyclicity, based on a connection with the theory of chase of TGDs in relational databases.

In our work, we have used the original notion of weak acyclicity. However, it is easy to adopt more advanced forms of acyclicity, since our results depend only on the ability of finding a finite bound on the number of distinct function terms that are generated (when applying the chase). While the majority of approaches that adopt forms of weak-acyclicity focus on databases (Marnette & Geerts, 2010; Meier, Schmidt, Wei, & Lausen, 2010), Cuenca Grau, Horrocks, Krötzsch, Kupke, Magka, Motik, and Wang (2012) investigate sophisticated forms of acyclicity in the context of knowledge bases without UNA. Their results can thus be seamlessly applied to KABs. Interestingly, to manage the impact of equalities in a setting without UNA, they resort to the *singularization* technique presented by Marnette (2009), which closely resembles the normalization of KABs introduced in Section 7.

Weak acyclicity allows us to gain decidability by bounding the number of distinct function terms that occur in the transition system. An alternative approach to gain decidability is to bound the number of distinct terms occurring in the ABox assertions of a state. Variants of this notion of "*state boundedness*" have been proposed recently in other contexts (Belardinelli et al., 2012; De Giacomo, Lesperance, & Patrizi, 2012; Bagheri Hariri et al., 2013). It is of great interest to explore such an approach in the setting presented here of actions acting on a description logic knowledge base.

We observe that our decidability result (as well as the ones commented here and in Section 8), comes with an algorithm for verification that is exponential in the size of the initial ABox. This precludes a direct application of these techniques to large-scale systems, without a careful analysis of how these can be modularized in small units to be verified (almost) separately. This is an important direction for further investigation.

**Acknowledgments**

This research has been partially supported by the EU under the ICT Collaborative Project ACSI (Artifact-Centric Service Interoperation), grant agreement n. FP7-257593, and under the large-scale integrating project (IP) Optique (Scalable End-user Access to Big Data), grant agreement n. FP7-318338.